\documentclass[journal]{IEEEtran}
\usepackage{amsmath,amsfonts}
\usepackage{algorithmic}
\usepackage{algorithm}
\usepackage{array}
\usepackage[caption=false,font=normalsize,labelfont=sf,textfont=sf]{subfig}
\usepackage{textcomp}
\usepackage{stfloats}
\usepackage{url}
\usepackage{verbatim}
\usepackage{graphicx}
\usepackage{balance}
\usepackage{cite}

\usepackage{booktabs}
\usepackage{multirow}
\usepackage{makecell}
\usepackage{threeparttable}

\usepackage{hyperref}
\usepackage[resetlabels]{multibib}
\newcites{supp}{Supplementary References}

\begin{document}

\title{HADAR-Based Thermal Infrared Hyperspectral Image Restoration}

\author{
    Cheng~Dai,
    Jiale~Lin,
    Bingxuan~Song,
    Yifei~Chen,
    Jiashuo~Chen,
    Xin~Yuan,~\IEEEmembership{Senior Member,~IEEE},
    and Fanglin~Bao
    \thanks{The code is available at \url{https://github.com/jialelin2007/HAIR}.}
    \thanks{Cheng Dai, Yifei Chen, and Fanglin Bao are with the School of Science, Westlake University, Hangzhou 310030, China.}
    \thanks{Jiale Lin, Bingxuan Song, Jiashuo Chen and Xin Yuan are with the School of Engineering, Westlake University, Hangzhou 310030, China.}
    \thanks{E-mails: \{daicheng, linjiale, songbingxuan, chenyifei99, chenjiashuo, xyuan, baofanglin\}@westlake.edu.cn.}
    \thanks{Cheng Dai and Jiale Lin contributed equally to this work.}
    \thanks{Corresponding author: Fanglin Bao.}
}

\markboth{HADAR-Based Thermal Infrared Hyperspectral Image Restoration}
{C. Dai \MakeLowercase{\textit{et al.}}: HADAR-Based Thermal Infrared Hyperspectral Image Restoration}

\maketitle

\begin{abstract}
    Thermal-infrared (TIR) hyperspectral imagery (HSI) provides critical scene information for various applications. However, its practical utility is severely limited by unique sensor degradations beyond the capabilities of existing restoration methods, which are ignorant of underlying thermal physics. Here, we propose HAIR (HADAR-based Image Restoration) as a physics-driven framework for ground-based TIR-HSI restoration. HAIR utilizes the HADAR rendering equation (HRE) and combines it with the atmospheric downwelling radiative transfer equation (RTE) to model TIR-HSI using temperature, emissivity, and texture (TeX) physical triplets. This physical model leads to a TeX decompose-synthesize strategy that guarantees physical consistency and spatio-spectral noise resilience, in stark contrast to existing approaches. Moreover, our framework uses a forward-modeled atmospheric downwelling reference, along with spectral smoothness of emissivity and blackbody radiation, to enable spectral calibration and generation that would otherwise be elusive. Our extensive experiments on the outdoor DARPA Invisible Headlights dataset and in-lab FTIR measurements show that HAIR consistently outperforms state-of-the-art methods across denoising, inpainting, spectral calibration, and spectral super-resolution, establishing a benchmark in objective accuracy and visual quality.
\end{abstract}


\section{INTRODUCTION}
\IEEEPARstart{T}{he} emerging HADAR (heat-assisted detection and ranging)~\cite{BaoHeatassistedDetectionRanging2023} based on thermal infrared (TIR) hyperspectral imagery (HSI) is driving a paradigm shift in physics-driven perception~\cite{xu2026universalcomputationalthermalimaging}. By highlighting the importance of the sub-leading scattering signal in the rendering equation and designing a proper model for it, HADAR recovers vivid geometric textures ($X$), in addition to temperature ($T$) and emissivity ($e$), from low-contrast ground-based TIR-HSI. This modified HADAR rendering equation (HRE)~\cite{BaoHeatassistedDetectionRanging2023, BaoWhyThermalImages2024} extends traditional temperature–emissivity (TE) separation to TeX decomposition. The resulting TeX vision, even at night, strikingly mimics RGB vision in daylight, exhibiting powerful night vision for autonomous driving, healthcare, and 3D modeling~\cite{DorkenGallastegiAbsorptionBasedPassiveRange2025, KushidaAffineTransformRepresentation2024, NGThermalVoyagerComparative2024, YellinConcurrentBandSelection2024, ZhangTADARThermalArraybased2024, HanHyperspectralPhasorThermography2025, YeThermalNeRFNeuralRadiance2024, XuColorRouterbasedLongwave2024}. However, HADAR's applicability relies fundamentally on the spatio-spectral fidelity of TIR-HSI, whereas real-world TIR-HSI acquisitions are inevitably corrupted by sensor noise, dead bands, spectral undersampling, and wavelength shifts. These artifacts not only pollute the visual perception of HSI but also disrupt the thermodynamic structures. TIR-HSI restoration is therefore urgent for HADAR-related applications.

According to the HRE, ground-based TIR-HSI is a superposition of target thermal emission and scattered environmental radiation (see Fig.~A2 in the Supplementary Appendix for more details). The observed radiance is therefore governed by the latent TeX attributes rather than a single latent image, as is commonly assumed in conventional restoration methods~\cite{ZhuangFastHyperspectralImage2018, LuTensorRobustPrincipal2020, TRLRF_AAAI2019, NGMeet_hwwei_tpami_2022, DIP2D-DIP3D-ICCVW2019, DDS2M_2023_ICCV}. Note that near the TeX degeneracy condition~\cite{BaoHeatassistedDetectionRanging2023,xu2026universalcomputationalthermalimaging}, even slight radiance-domain modifications that appear beneficial in HSI space may lead to incorrect TeX decompositions. Consequently, TIR-HSI restoration must preserve physical consistency alongside visual quality.

To utilize the TeX model for physics-consistent restoration, sensor-aware characterization of TIR-HSI degradations is crucial. Pushbroom camera suffers from stripe artifacts caused by focal-plane non-uniformity and flicker noise~\cite{nuc_priciple_causes_1988, NUCproblem_TV_fix_algorithm_2010}. More importantly, both pushbroom and Fourier Transform Infrared (FTIR) cameras share band-dependent thermal noise, spectral undersampling, non-uniform wavelength shifts, and catastrophic band-wise corruption~\cite{HgCdTe_noise_nonstable2002, FTIR_noise_analysis_optical_engineering_2012noise, FTIR_spectral_shift_2006optical, hsi_smile_effect_2010, YellinConcurrentBandSelection2024} that require an external anchor for restoration. LibRadtran~\cite{libradtran_software_emde2016libradtran} is commonly used for that purpose to generate the atmospheric downwelling reference. Specifically, LibRadtran solves the downwelling radiative transfer equation (RTE) from measurable atmospheric profiles and molecular spectroscopy, to model the high-resolution atmospheric downwelling spectra~\cite{GORDON2026109807} (see Supplementary Appendix~B.1).

In this paper, we present HAIR (Fig.~\ref{fig:main_structure}), a HADAR-based image restoration framework designed to systematically resolve TIR-HSI denoising, inpainting, spectral calibration, and spectral super-resolution across diverse detector architectures.

Our main contributions are:
\begin{itemize}
    \item First, we propose and demonstrate a physics-driven TeX decompose-synthesize strategy for physics-consistent TIR-HSI restoration and reliable temperature, emissivity, and texture decomposition.

    \item Second, we formulate a downwelling-guided calibration scheme by aligning observed atmospheric signatures with an RTE-simulated reference. This scheme estimates spectral shifts and enables band calibration, completion, and enhancement while preserving thermodynamic structure.

    \item Third, we analyze sensor degradation mechanisms in pushbroom and FTIR TIR-HSI, and propose a unified denoising model that separates tractable perturbations from invalid spectral measurements.

    \item Fourth, we experimentally test HAIR on denoising, inpainting, spectral calibration, and spectral super-resolution, consistently demonstrating that HAIR outperforms existing methods in objective accuracy, visual quality, and thermodynamic consistency, with practical robustness and scalability across sensor architectures.
\end{itemize}

\section{RELATED WORK}
\subsection{TIR-Specific Methods}
Most TIR restoration methods target airborne or satellite observations and are built on long-range atmospheric RTE assumptions. BBSTV~\cite{BBSTV_2021} uses an RTE-based bidirectional framework for dead-line removal and degraded-band restoration, while SNRSWAC~\cite{wangdu_2025_igarss} and WBAC-TES~\cite{clq_20260301_atmpspheric_correction} focus on atmospheric correction for UAV and satellite imagery. These assumptions differ from ground-based TIR-HSI, where short-range interactions and downwelling radiance dominate image formation, making such formulations difficult to transfer.

\subsection{General Model-Based Methods}
General model-based HSI restoration mainly relies on VIS--NIR priors, including global low-rank subspaces~\cite{HySime_TGRS_2008, ZhuangFastHyperspectralImage2018, LRMR_TGRS_Zhanghongyan_Hewei_2013, wnnm_cvpr2014, LuTensorRobustPrincipal2020, TRLRF_AAAI2019}, non-local self-similarity~\cite{BM3D_2007_TIP, BM4D_TIP_2012, NLSS_2018, NGMeet_hwwei_tpami_2022}, and tensor/TV regularization~\cite{SSTV_2016, LRTDTV_2019, E3DTV_TIP2020}. Although effective for reflective imaging, these priors do not match the emissive formation of TIR-HSI: GLR and NSS may suppress high-frequency atmospheric signatures needed for atmosphere-related applications~\cite{XuColorRouterbasedLongwave2024}, while TV-type priors may introduce oversmoothing or staircase artifacts~\cite{TGV_2010_staircasing_effect}. They also do not explicitly address spectral shifts. Existing VIS--NIR spectral super-resolution (SSR) methods usually assume paired low-resolution HSI and high-resolution RGB/PAN/MSI observations~\cite{RGB_HSI_SR_ICCV2016, MSI_HSI_SR_TIP2021, MSI_HSI_SR_tpami2025}, an assumption rarely satisfied in ground-based TIR imaging~\cite{YellinConcurrentBandSelection2024}.

\subsection{Deep Learning-Based Methods}
Deep learning methods replace handcrafted priors with learned representations, including CNN/recurrent/Transformer models~\cite{AWAN_CVPRW2020, QRNN3D_TNN2021, HSRNet_TNN2022, DRCRNet_cvprw2022, LTRN_TNN2025, SSRT_UNet_TGRS2024, SST_AAAI2023, MSTpp_cvpr2022}, deep unfolding networks~\cite{HLRTF_CVPR2022, MACNet_deep_unfolding_TGRS2022, FlexDLD_TIP2024}, and unsupervised or self-supervised frameworks~\cite{DIP2D-DIP3D-ICCVW2019, DDS2M_2023_ICCV, Diff_unmix_cvpr2024}. However, these methods are mostly designed for generic or VIS--NIR HSI restoration and do not explicitly encode TIR radiative transfer or sensor-specific degradation. Optimized mainly by statistical reconstruction losses, they may produce visually plausible spectra while attenuating atmospheric signatures and weakening the thermodynamic consistency required by HRE-constrained HADAR inversion~\cite{BaoHeatassistedDetectionRanging2023,xu2026universalcomputationalthermalimaging}.

\section{DEGRADATION ANALYSIS AND MODELING}
\label{sec:degradation_model}
\subsection{Unified Sensor-Aware Degradation Model}
Ground-based TIR-HSI is mainly acquired by pushbroom cameras or FTIR spectrometers~\cite{YellinConcurrentBandSelection2024, BaoHeatassistedDetectionRanging2023, DorkenGallastegiAbsorptionBasedPassiveRange2025}, which share band-dependent stochastic noise, spectral undersampling, wavelength shifts, and catastrophic band failures~\cite{FTIR_noise_analysis_optical_engineering_2012noise, FTIR_spectral_shift_2006optical, thermal_gaussian_noise_rs2020_ftir, BaoHeatassistedDetectionRanging2023}; pushbroom systems further exhibit detector-nonuniformity stripes~\cite{HgCdTe_noise_nonstable2002, nuc_priciple_causes_1988, 2point_num_correction_2019, nuc-correction_noise_analysis_1995, BaoHeatassistedDetectionRanging2023, radiation_calibration_fix_2001_nuc}. We model these effects as:

\begin{equation}
    \label{eq:unified_model}
    \mathcal{Y} = \mathcal{M} \odot (\mathcal{H}(\mathcal{X}) + \mathcal{S} + \mathcal{N}) + (\mathcal{I} - \mathcal{M}) \odot \mathcal{C},
\end{equation}
where $\mathcal{X}$ is the latent pristine HSI, $\mathcal{I}$ is an all-ones tensor, and $\odot$ denotes element-wise multiplication. The spectral operator $\mathcal{H}$ models undersampling and wavelength shifts; $\mathcal{S}$ represents tractable structured perturbations; $\mathcal{N}$ denotes stochastic noise; and the mask $\mathcal{M}$ with component $\mathcal{C}$ separates valid observations from invalid measurements caused by catastrophic failure. This decomposition distinguishes recoverable perturbations from measurements that no longer preserve information.

\subsection{Tractable Spectral and Spatial Perturbations}
For valid bands, the shared stochastic component can be approximated as band-dependent Gaussian noise:

\begin{equation}
    \label{eq:gaussian_noise}
    \mathcal{N}(i, j, k) \sim \mathcal{N}(0, \sigma_k^2),
\end{equation}
where $\sigma_k^2$ is the variance of the $k$-th band. The band dependence reflects row-dependent sensitivity in pushbroom sensors~\cite{pushbroom_fpa_science1985} and wave-number-dependent Noise Equivalent Spectral Radiance (NESR) in FTIR systems~\cite{FTIR_noise_analysis_optical_engineering_2012noise}. Finite detector bandwidth and central-wavelength instability are captured by the shift-aware spectral acquisition model

\begin{equation}
    \label{eq:spectral_undersampling}
    \mathcal{H}(\mathcal{X})(i, j, k)
    = \sum_{k'=1}^{C} f_{\lambda_{k}}(\lambda_{k'}; \Delta\lambda_{k}, \theta_{k}) \mathcal{X}(i, j, k'),
\end{equation}
for $k = 1,\ldots,c$, where $\mathcal{X} \in \mathbb{R}^{H \times W \times C}$ is the latent high-resolution HSI, $\mathcal{H}(\mathcal{X}) \in \mathbb{R}^{H \times W \times c}$ ($c \ll C$) is the ideal clean observation, and $f_{\lambda_k}$ is the discrete spectral response function (SRF) with shift $\Delta\lambda_k$ and shape parameter $\theta_k$. Energy conservation is enforced by $\sum_{k'=1}^{C} f_{\lambda_{k}}(\lambda_{k'}; \Delta\lambda_{k}, \theta_{k}) = 1$. We instantiate this SRF as

\begin{equation}
    \label{eq:srf_gaussian_shift}
    f_{\lambda_{k}}(\lambda_{k'}; \Delta\lambda_{k}, \sigma)
    =
    \frac{\exp\left( -\frac{(\lambda_{k'} - (\lambda_{k} + \Delta\lambda_{k}))^2}{2\sigma^2} \right)}
    {\sum_{t=1}^{C} \exp\left( -\frac{(\lambda_t - (\lambda_{k} + \Delta\lambda_{k}))^2}{2\sigma^2} \right)},
\end{equation}
where $\sigma$ controls the effective bandwidth.

Pushbroom acquisition further introduces tractable stripe artifacts because focal-plane non-uniformity is replicated along the scan direction~\cite{HgCdTe_noise_nonstable2002, nuc_priciple_causes_1988, nuc-correction_noise_analysis_1995, BaoHeatassistedDetectionRanging2023}. Following established models~\cite{nuc_linear_scientific_report_2021, NUCproblem_TV_fix_algorithm_2010, nuc_linear_solution_optics_2023, 2point_num_correction_2019, dl_nuc_correction_2022, radiation_calibration_fix_2001_nuc}, we use a gain-bias form:
\begin{equation}
    \label{eq:stripe_noise_components}
    \mathcal{S}(i, j, k) =
    \mathcal{A}(i, k) \cdot \mathcal{H}(\mathcal{X})(i, j, k) + \mathcal{B}(i, k),
\end{equation}
where $i$, $j$, and $k$ denote cross-track detector index, along-track position, and spectral band, and $\mathcal{A}(i,k)$ and $\mathcal{B}(i,k)$ are detector-specific gain and bias.

\subsection{Invalid Measurements and Restoration Implications}
Catastrophic corruption invalidates whole bands, e.g., from pushbroom detector failures or FTIR edge-band signal collapse~\cite{pushbroom_fpa_science1985, BaoHeatassistedDetectionRanging2023, FTIR_noise_analysis_optical_engineering_2012noise, thermal_gaussian_noise_rs2020_ftir}. We mark valid samples by $\mathcal{M} \in \{0,1\}^{H \times W \times c}$; for a failed band set $\Omega_d$, $\mathcal{M}(:, :, \Omega_d)=\mathbf{0}$ and

\begin{equation}
    \label{eq:catastrophic_corruption}
    \mathcal{Y}_{\text{catastrophic}} = (\mathcal{I} - \mathcal{M}) \odot \mathcal{C},
\end{equation}
where $\mathcal{C} \in \mathbb{R}^{H \times W \times c}$ denotes arbitrary non-physical values recorded under hardware failure.

Thus HAIR separates tractable perturbations from invalid measurements: noise, shifts, and stripes are restored or calibrated on valid bands, while catastrophic bands are excluded and recovered using external physical guidance.

\subsection{Image-Restoration Objective}
Given a degraded HSI $\mathcal{Y}$, we recover a radiometrically clean and physically interpretable target $\mathcal{Y}_{\mathrm{tar}}$ on $\Lambda_{\mathrm{tar}}=\{\lambda'_k\}_{k=1}^{c'}$ by applying the SRF in Eq.~\ref{eq:srf_gaussian_shift} to the latent pristine $\mathcal{X}$:
\begin{equation}
    \label{eq:target_hsi_relation}
    \mathcal{Y}_{\mathrm{tar}}(i,j,k)
    =
    \sum_{k'=1}^{C}
    f_{\lambda'_k}(\lambda_{k'}; 0, \theta'_k)
    \mathcal{X}(i,j,k'),
\end{equation}
where $\mathcal{Y}_{\mathrm{tar}}\in\mathbb{R}^{H\times W\times c'}$, $f_{\lambda'_k}$ is centered at $\lambda'_k$, and $\theta'_k$ controls the target response shape. This objective unifies denoising, inpainting, spectral calibration, and spectral super-resolution.

\begin{figure*}[t]
    \centering
    \includegraphics[width=\linewidth]{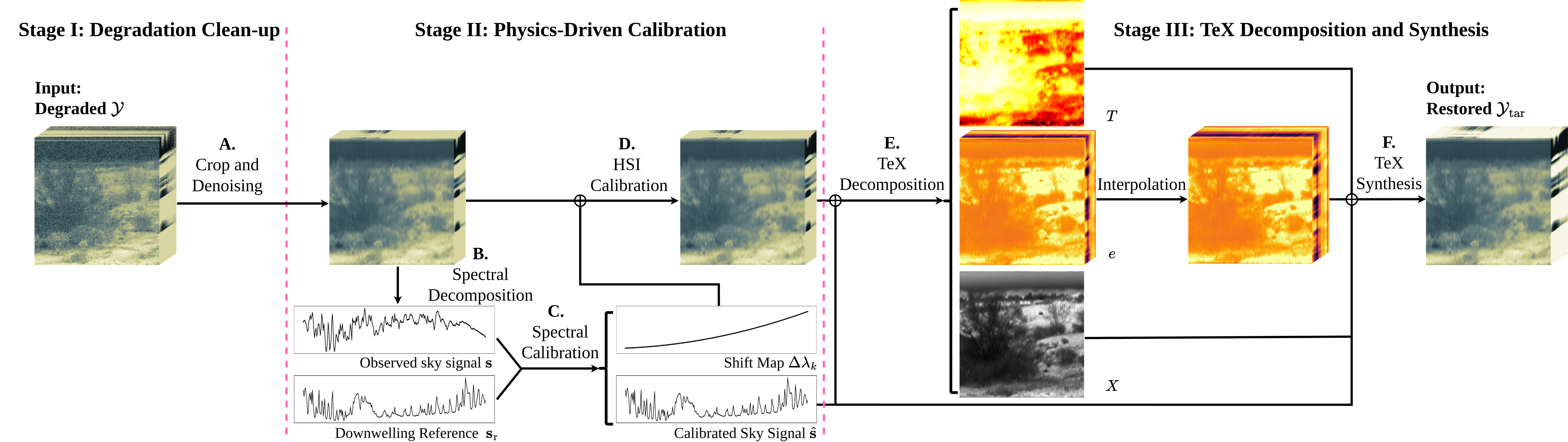}
    \caption{Flowchart of the HAIR framework. Given a degraded HSI $\mathcal{Y}$, HAIR first extracts the observed sky signal and uses a simulated downwelling reference for spectral calibration (A-D). Then, HAIR uses the core TeX decompose-synthesize mechanism (E-F) to reconstruct a physics-consistent, noise-resilient $\mathcal{Y}_{\mathrm{tar}}$.}
    \label{fig:main_structure}
\end{figure*}

\section{PROPOSED METHOD}
\label{sec:method}
\subsection{The Overall HAIR Framework}
Building on Section~\ref{sec:degradation_model}, HAIR restores $\mathcal{Y}_{\mathrm{tar}}$ through three HRE-consistent stages (Fig.~\ref{fig:main_structure}): architecture-aware clean-up, downwelling-guided wavelength calibration (Supplementary Appendix~B.1), and TeX decomposition-synthesis. Algorithm~\ref{alg:hair_unified} summarizes the three-stage execution for denoising, inpainting, spectral calibration, and spectral super-resolution. The mathematical formulations are given below.

\subsection{Stage I: Degradation Clean-up}
\label{sec:stage_1_cleanup}

Stage I institutes a hierarchical, sensor-aware clean-up protocol, including catastrophic band detection, adaptive variational destriping, and subspace-based denoising. By excluding invalid bands and suppressing tractable spatial degradations on the valid spectrum, it establishes a reliable radiance input for subsequent spectral calibration and HADAR inversion.

\subsubsection{Automated Catastrophic Band Exclusion (Module A.I)}

Catastrophically corrupted bands are detected before spatial restoration using two architecture-aware scores. The stochastic noise score $s_{1,k}$ is estimated via HySime~\cite{HySime_TGRS_2008}; bands with $s_{1,k}>\tau_1$ (e.g., $\tau_1=0.01$) are treated as thermally compromised. For pushbroom data, a stripe score $s_{2,k}$ is computed from the cross-track row-mean profile $\mathbf{r}_k \in \mathbb{R}^H$. With $\tilde{\mathbf{r}}_k=\mathbf{r}_k\ast g_\sigma$ denoting a Gaussian-smoothed baseline (e.g., $\sigma=10.0$), the stripe strength is
\begin{equation}
    \label{eq:stripe_strength}
    s_{2,k} = \sqrt{ \frac{1}{H} \sum_{i=1}^{H} (r_{i,k} - \tilde{r}_{i,k})^2 }.
\end{equation}
Pushbroom bands exhibiting $s_{2,k} > \tau_2$ (e.g., $\tau_2 = 0.03$) are flagged for severe directional corruption.

The initial catastrophic-band set $\Omega_{\text{c}}$ is selected by sensor architecture: FTIR uses only the stochastic score, while pushbroom (PB) uses either degradation score:

\begin{equation}
    \label{eq:candidate_set}
    \Omega_{\text{c}} = \left\{ k \in \{1, \dots, c\} \;\middle|\;
    \begin{cases}
        s_{1,k} > \tau_1,                       & \text{FTIR} \\
        s_{1,k} > \tau_1 \lor s_{2,k} > \tau_2, & \text{PB}
    \end{cases}
    \right\}.
\end{equation}

To avoid over-masking and preserve spectral continuity, the discarded ratio is capped by $\gamma_{\text{max}}$ (e.g., $0.3$). Candidates beyond this budget are ranked by $s_k=s_{1,k}$ for FTIR and $s_k=s_{1,k}+s_{2,k}$ for pushbroom. Let $\mathcal{T}(\Omega,s,K)$ return the $K$ largest-score indices; the final dead-band set is
\begin{equation}
    \label{eq:dead_band_set}
    \Omega_{\text{d}} =
    \begin{cases}
        \Omega_{\text{c}},                                                                         & \text{if } |\Omega_{\text{c}}| \le \lfloor \gamma_{\text{max}} c \rfloor \\
        \mathcal{T}\left(\Omega_{\text{c}}, \{s_k\}, \lfloor \gamma_{\text{max}} c \rfloor\right), & \text{otherwise}
    \end{cases}.
\end{equation}

The valid set is $\Omega_{\text{g}} = \{1, \dots, c\} \setminus \Omega_{\text{d}}$, and the tractable radiance tensor is sliced as
\begin{equation}
    \label{eq:cropped_hsi}
    \mathcal{Y}(:, :, \Omega_{\text{g}}) \in \mathbb{R}^{H \times W \times |\Omega_{\text{g}}|},
\end{equation}
\noindent where $\mathcal{Y}$ is the original degraded measurement.

\subsubsection{Adaptive Variational Destriping (Module A.II)}
\label{sec:destriper_adv_moduleA.II.}
For valid pushbroom bands, $\mathcal{Y}(:, :, \Omega_{\text{g}})$ is decomposed into a non-stripe component $\mathcal{Z}$ and a stripe component $\mathcal{S}$, while residual stochastic noise is left for Module A.III. We minimize
\begin{equation}
    \label{eq:destriper_main}
    \min_{\mathcal{Z}, \mathcal{S}}
    \frac{1}{2}\|\mathcal{Y} - \mathcal{Z} - \mathcal{S}\|_F^2
    + \mathcal{R}_{1}(\mathcal{Z}) + \mathcal{R}_{2}(\mathcal{S}).
\end{equation}

To preserve sharp structures while reducing staircase artifacts~\cite{TGV_2010_staircasing_effect}, $\mathcal{R}_{1}$ combines first- and second-order TV:
\begin{equation}
    \label{eq:destriper_hsi_regularization}
    \mathcal{R}_{1}(\mathcal{Z}) = \lambda_1 \|\nabla_x \mathcal{Z}\|_1 + \alpha(s_{2,k}) \|\nabla_y \mathcal{Z}\|_1 + \lambda_3 \|\nabla_{yy}\mathcal{Z}\|_1,
\end{equation}
where $\alpha(s_{2,k}) = m s_{2,k}$ is an adaptive vertical-smoothing weight with fixed scale $m$ (e.g., $m=2$).

To capture the directional sparsity of stripe artifacts, $\mathcal{R}_{2}$ penalizes horizontal gradients and overall sparsity:
\begin{equation}
    \label{eq:destriper_stripe_regularization}
    \mathcal{R}_2(\mathcal{S}) = \lambda_4 \|\nabla_x \mathcal{S}\|_1 + \lambda_5 \|\mathcal{S}\|_1.
\end{equation}

The optimization is solved by ADMM, with detailed derivations and empirical convergence validation in Supplementary Appendix~A.

\subsubsection{Subspace Denoising (Module A.III)}
To suppress remaining stochastic noise $\mathcal{N}$, we use FHyDe~\cite{ZhuangFastHyperspectralImage2018} with the noise estimates $s_{1,k}$ from Module A.I. Its subspace projection and BM3D-based non-local filtering~\cite{BM3D_2007_TIP} denoise the HSI while preserving atmospheric signatures needed for calibration.

\subsection{Stage II: Physics-Driven Calibration}
Stage II constructs physical spectral priors from the cleaned valid-band radiance: Module B extracts the sky signal, Module C aligns it with a forward-modeled reference, and Module D propagates the calibrated wavelength to the valid HSI.

\subsubsection{Observed Sky Signal Extraction (Module B)}
\label{sec:atmospheric_signature_extraction}
We first average the cleaned valid-band HSI over all pixels to obtain a scene-level spectrum $\mathbf{y}\in\mathbb{R}^{|\Omega_g|}$. Given $\mathbf{y}$, the atmospheric signature is extracted by asymmetric least-squares (ALS) baseline separation~\cite{ALS_2025}. $\mathbf{y}$ is decomposed into a smooth thermal baseline $\mathbf{b}$ and a high-frequency component $\mathbf{s}'$ by solving
\begin{equation}
    \label{eq:als_baseline}
    \min_{\mathbf{b}} \sum_{i=1}^{|\Omega_g|} \omega_i (y_i-b_i)^2
    + \beta \sum_{i=3}^{|\Omega_g|} (\Delta^2 b_i)^2
\end{equation}
where $\Delta^2 b_i = b_i - 2b_{i-1} + b_{i-2}$, $\beta$ controls baseline smoothness (e.g., $10^4$), and $\omega_i$ are asymmetric weights updated iteratively. The atmospheric signature is $\mathbf{s}'=\mathbf{y}-\mathbf{b}$.

We then interpolate $\mathbf{s}'$ onto the full sensor index set:

\begin{equation}
    \label{eq:observed_sky_interpolation}
    \mathbf{s} = \mathcal{Q}\!\left(\mathbf{s}', \Omega_g \rightarrow \{1,\dots,c\}\right) \in \mathbb{R}^{c},
\end{equation}
where $\mathcal{Q}$ denotes a 1D interpolation operator and $\mathbf{s}(k)$ is indexed by the sensor band index $k \in \{1,\dots,c\}$.

\subsubsection{Spectral Calibration and Band Reassignment (Modules C, D)}
\label{sec:spectral_calibration}

We forward-model a high-resolution downwelling reference $\mathbf{s}_{\mathrm{r}}$ via libRadtran, as detailed in Supplementary Appendix~B.1.

Calibration distinguishes nominal wavelengths $\{\lambda_k\}_{k=1}^{c}$ from actual operating wavelengths $\{\lambda_k^*\}_{k=1}^{c}$, parameterized as
\begin{equation}
    \label{eq:actual_wavelength}
    \lambda_k^* = a k^2 + b k + d,
    \quad k = 1,\ldots,c.
\end{equation}

Given $\lambda_k^*$, $\mathbf{s}_{\mathrm{r}}$ is projected to the sensor domain through the SRF $f_{\lambda_k}(\lambda_{k'};\lambda_k^*-\lambda_k,\sigma)$ in Eq.~\ref{eq:srf_gaussian_shift}, with Z-score normalization matching the observed amplitude:
\begin{equation}
    \label{eq:calibrated_sky}
    \hat{\mathbf{s}}(k)
    =
    \mu_{\mathbf{s}}+\sigma_{\mathbf{s}}\mathcal{Z}\left(\sum_{k'=1}^{C} \mathbf{s}_{\mathrm{r}}(k')\,
    f_{\lambda_k} (\lambda_{k'}; \lambda_k^*-\lambda_k, \sigma)\right)
\end{equation}
for $k = 1,\ldots,c$, where $\hat{\mathbf{s}}$ is the simulated sensor-domain spectrum and $\mu_{\mathbf{s}},\sigma_{\mathbf{s}}$ are the mean and standard deviation of $\mathbf{s}$. The calibration parameters are estimated by
\begin{equation}
    \label{eq:spectral_correction}
    \min_{\sigma,a,b,d}
    \left\|
    \hat{\mathbf{s}}
    -
    \mathbf{s}
    \right\|_2^2.
\end{equation}
This nonlinear problem is solved by the parallel grid search, yielding the shift map
\begin{equation}
    \label{eq:shift_map}
    \Delta\lambda_k = \lambda_k^* - \lambda_k,
    \quad k = 1,\ldots,c.
\end{equation}

Instead of resampling radiance to the nominal grid, we re-associate valid bands with calibrated operating wavelengths:
\begin{equation}
    \label{eq:calibrated_hsi}
    \mathcal{Y}_{\mathrm{c}}(x,y,\lambda_k^*)
    =
    \mathcal{Y}_{\mathrm{denoised}}(x,y,k), \quad k \in \Omega_g
\end{equation}
where $\mathcal{Y}_{\mathrm{denoised}}$ is the Stage~I output. Stage~II thus produces $\hat{\mathbf{s}}$ and $\mathcal{Y}_{\mathrm{c}}$ for TeX decomposition and synthesis.

\subsection{Stage III: TeX Decomposition and Synthesis}

Stage III decomposes $\mathcal{Y}_{\mathrm{c}}$ into TeX factors and synthesizes the restored HSI $\mathcal{Y}_{\mathrm{tar}}\in\mathbb{R}^{H\times W\times c'}$ on the target grid $\Lambda_{\mathrm{tar}}=\{\lambda'_k\}_{k=1}^{c'}$. This grid is the nominal sensor grid for denoising, inpainting, and spectral calibration, and the desired finer grid for spectral super-resolution.

\subsubsection{HADAR-Based TeX Decomposition (Module E)}
HADAR inversion uses $\mathcal{Y}_{\mathrm{c}}$ and $\hat{\mathbf{s}}$ on $\Lambda^{*}=\{\lambda_k^{*}\}_{k\in\Omega_g}$. Omitting pixel indices, the HRE~\cite{BaoHeatassistedDetectionRanging2023, xu2026universalcomputationalthermalimaging} is
\begin{equation}
    \label{eq:hadar_rtm_pixel}
    S = eB(T)+\bigl(1-e\bigr)X
\end{equation}
where $e$ is emissivity, $B$ is blackbody radiation, and $X=\sum_{j\neq i}V_jS_j$ denotes texture from environment. HADAR solves
\begin{equation}
    \label{eq:hadar_inversion}
    \begin{aligned}
        \min_{e,T,\{V\}} \;
        \sum_{k\in\Omega_g}
        \Big(
        \mathcal{Y}_{\mathrm{c}}(\lambda_k^*)
        -
        \Big[
        e(\lambda_k^*)\,B(T,\lambda_k^*) \\
            \quad + \bigl(1-e(\lambda_k^*)\bigr)
            \sum_{j\neq i} V_j\,S_j(\lambda_k^*)
            \Big]
        \Big)^2,
    \end{aligned}
\end{equation}
disentangling the mixed radiance into $(e,T,V)$ for the following target-band synthesis.

\subsubsection{Target-Grid Projection (Module F.I)}
Emissivity is interpolated from valid calibrated bands to the target grid for its smooth spectral nature:
\begin{equation}
    \label{eq:interp_emissivity}
    e_{\mathrm{tar}} = \mathcal{Q}\!\left(e,\,\{\lambda_k^*\}_{k\in\Omega_g}\rightarrow\{\lambda_k'\}_{k=1}^{c'}\right),
\end{equation}
where $\mathcal{Q}$ is a 1D interpolation operator. The blackbody radiance is then evaluated on the target grid by Planck's law:
\begin{equation}
    \label{eq:planck_nominal}
    B(T,\lambda)
    =
    \frac{10^{24}\,2hc^2}{\lambda^{5}}
    \frac{1}{\exp\!\left(\frac{10^{6}hc}{\lambda k_B T}\right)-1}
\end{equation}
where $h$, $c$, and $k_B$ denote the Planck constant, speed of light, and Boltzmann constant, with $\lambda$ in micrometers and output in $\mathrm{W\cdot m^{-2}\cdot sr^{-1}\cdot \mu m^{-1}}$.

The downwelling sky signal is projected to the same grid using the SRF width $\sigma$ estimated in Eq.~\ref{eq:spectral_correction}:
\begin{equation}
    \label{eq:sky_nominal}
    \tilde{\mathbf{s}}(k)=
    \mu_{\mathbf{s}}+\sigma_{\mathbf{s}}\,
    \mathcal{Z}\left( \sum_{k'=1}^{c'} \mathbf{s}_{\mathrm{r}}\left(k'\right)f_{\lambda'_{k}}\left( \lambda_{k'};0,\sigma \right) \right)
\end{equation}
Thus $e(\lambda'_k)$, $B(T,\lambda'_k)$, and $\tilde{\mathbf{s}}(k)$ are on the same target grid.

\subsubsection{HRE Synthesis (Module F.II)}
We reconstruct texture $X$ through recursive radiative transfer between the target and its surroundings. Initialization assumes no reflection and includes downwelling sky radiance and direct thermal emission:
\begin{equation}
    \label{eq:init_texture}
    X^{(0)} = V_{\mathrm{sky}}\tilde{\mathbf{s}} + \sum_{j\neq i}V_j\,\mathbb{E}_{\Omega_j}\!\left[e\,B(T)\right],
\end{equation}
where $V_j$ and $V_{\mathrm{sky}}$ are view factors and $\mathbb{E}_{\Omega_j}[\cdot]$ averages over object region $\Omega_j$.

Texture is updated for $N$ bounces (typically $N=4$):
\begin{equation}
    \label{eq:forward_texture}
    X^{(n+1)} = X^{(0)} + \sum_{j\neq i}V_j\,\mathbb{E}_{\Omega_j}\!\left[(1-e)\,X^{(n)}\right],
\end{equation}
After convergence, $X=X^{(N)}$, and the restored HSI $\mathcal{Y}_{\mathrm{tar}}$ is synthesized via Eq.~\ref{eq:hadar_rtm_pixel}.

\begin{algorithm}[htbp]
    \caption{Unified HAIR Flow}
    \label{alg:hair_unified}
    \small
    \begin{algorithmic}[1]
        \REQUIRE Degraded HSI $\mathcal{Y}$, camera type $\xi\in\{\mathrm{PB},\mathrm{FTIR}\}$, wavelength grids $\Lambda$ and $\Lambda_{\mathrm{tar}}$, and atmospheric parameters.
        \ENSURE Restored HSI $\mathcal{Y}_{\mathrm{tar}}$.

        \STATE \textbf{Stage I: Degradation Clean-up}
        \STATE Estimate $\{s_{1,k}\}$ via HySime~\cite{HySime_TGRS_2008}.
        \IF{$\xi=\mathrm{PB}$}
        \STATE Estimate $\{s_{2,k}\}$ via Eq.~\ref{eq:stripe_strength}.
        \ENDIF
        \STATE Determine $\Omega_g$ via Eqs.~\ref{eq:candidate_set} and~\ref{eq:dead_band_set} and crop valid-band HSI.
        \IF{$\xi=\mathrm{PB}$}
        \STATE Remove stripe artifacts on $\Omega_g$ by solving Eq.~\ref{eq:destriper_main}.
        \ENDIF
        \STATE Remove stochastic noise via FHyDe~\cite{ZhuangFastHyperspectralImage2018}.

        \STATE \textbf{Stage II: Physics-Driven Calibration}
        \STATE Extract $\mathbf{s}$ via Eqs.~\ref{eq:als_baseline} and~\ref{eq:observed_sky_interpolation} from $\mathcal{Y}_{\mathrm{denoised}}(:,:,\Omega_g)$.
        \STATE Obtain $\mathbf{s}_{\mathrm{r}}$ with libRadtran (Supplementary Appendix~B.1).
        \STATE Estimate $\lambda_k^*$ and $\sigma$ via Eqs.~\ref{eq:actual_wavelength} and~\ref{eq:spectral_correction}.
        \STATE Compute $\Delta\lambda_k$ and $\hat{\mathbf{s}}$ via Eqs.~\ref{eq:shift_map} and~\ref{eq:calibrated_sky}.
        \STATE Re-associate $\mathcal{Y}_{\mathrm{denoised}}$ with $\lambda_k^*$ via Eq.~\ref{eq:calibrated_hsi}.

        \STATE \textbf{Stage III: TeX Decomposition and Synthesis}
        \STATE Recover $e$, $T$, and $V$ via HADAR inversion in Eq.~\ref{eq:hadar_inversion}.
        \STATE Interpolate $e$ to $\Lambda_{\mathrm{tar}}$ via Eq.~\ref{eq:interp_emissivity}.
        \STATE Generate thermal radiance $B(T,\lambda)$ via Eq.~\ref{eq:planck_nominal}.
        \STATE Project $\mathbf{s}_{\mathrm{r}}$ onto $\Lambda_{\mathrm{tar}}$ via Eq.~\ref{eq:sky_nominal}.
        \STATE Compute texture $X$ via Eqs.~\ref{eq:init_texture} and~\ref{eq:forward_texture}.
        \STATE Reconstruct $\mathcal{Y}_{\mathrm{tar}}$ via HRE in Eq.~\ref{eq:hadar_rtm_pixel}.
        \RETURN $\mathcal{Y}_{\mathrm{tar}}$
    \end{algorithmic}
\end{algorithm}

\section{EXPERIMENTS}
\label{sec:experiments}
We evaluate HAIR on benchmark and real-world TIR-HSI data across denoising, inpainting, spectral calibration, and spectral super-resolution, measuring objective accuracy, visual quality, and thermodynamic consistency.

\subsection{Experimental Datasets}
We use two ground-based TIR hyperspectral datasets covering pushbroom and FTIR architectures. Both are acquired from lateral terrestrial viewpoints, with radiance approximately in the range of $4$--$15~\mathrm{W\cdot m^{-2}\cdot sr^{-1}\cdot \mu m^{-1}}$.

\subsubsection{Pushbroom}
We use the TIR subset of the DARPA dataset~\cite{YellinConcurrentBandSelection2024}, which contains natural and urban scenes acquired under varying conditions. The cubes have approximately 250--256 bands over $8$--$13 \mu$m. Deep learning baselines are trained on 100 cubes, and quantitative evaluation uses 10 scene-disjoint cubes (IH-10) with synthetic degradations.

\subsubsection{FTIR}
We use in-lab data collected at Wuhan University with a Telops Hyper-Cam LW system. The dataset covers materials and scenes such as plastics, gypsum, buildings, and grassland; each cube has size $256\times320\times86$ over $7.88$--$11.48\,\mu$m. Deep learning baselines are trained on 50 cubes.

\subsection{Evaluation Protocol}
Quantitative evaluation on IH-10 uses PSNR, SSIM, ERGAS~\cite{ERGAS_2000}, RMSE, and SAM~\cite{SAM_1992} for denoising, inpainting, and spectral super-resolution. For real data, where pixel-wise HSI and TeX ground truth are unavailable, we evaluate visual quality and physical plausibility: emissivity should be regionally coherent and consistent with references such as the NASA spectral library~\cite{NASA_ECOSTRESS_SpectralLib_2019}, while temperature should remain smooth within homogeneous regions. This protocol follows HADAR's TeX-degeneracy analysis~\cite{BaoHeatassistedDetectionRanging2023, xu2026universalcomputationalthermalimaging}. For texture visualization, we set $V_j=0$ only at the display stage of Eqs.~\eqref{eq:init_texture}--\eqref{eq:forward_texture}, so the displayed texture is driven by the sky term.

\subsection{Simulated Experimental Setup}
All synthetic degradations follow Eq.~\ref{eq:unified_model}. Unless otherwise stated, Gaussian noise has variance $\sigma^2=0.5$, stripe noise follows Eq.~\ref{eq:stripe_noise_components} with $\mathcal{A} \sim \mathcal{N}(0,0.2^2)$ and $\mathcal{B} \sim \mathcal{N}(1,0.5^2)$, and the mask $\mathcal{M}$ uses stripe densities $\tilde{s}=0.05$ and $\tilde{s}=0.2$ for normal and corrupted bands with baseline corrupted ratio $\tilde{b}=0.1$. Task-specific settings are:

\begin{itemize}
    \item Denoising: We vary $\sigma^2 \in \{0.1, 0.5, 1.0\}$ and band stripe density $\tilde{s} \in \{0.01, 0.05, 0.1\}$, yielding 9 combinations.

    \item Inpainting: Corrupted bands use stronger degradations, $\mathcal{A} \sim \mathcal{N}(0,1.0^2)$, $\mathcal{B} \sim \mathcal{N}(4.0,0.5^2)$, and $\tilde{s}=0.5$, with $\tilde{b} \in \{0.2, 0.3, 0.5\}$.

    \item Spectral Super-Resolution (SSR): We apply Eq.~\ref{eq:srf_gaussian_shift} with $\sigma^2=1.0$ and evaluate $\times2$, $\times4$, and $\times8$ scales.
\end{itemize}

\subsection{TIR-HSI Denoising Experiments}
We compare HAIR with ten denoising baselines, including model-based methods SSTV~\cite{SSTV_2016}, LRTDTV~\cite{LRTDTV_2019}, E3DTV~\cite{E3DTV_TIP2020}, NGMeet~\cite{NGMeet_hwwei_tpami_2022}, and FHyDe~\cite{ZhuangFastHyperspectralImage2018}, and deep models DIP2d~\cite{DIP2D-DIP3D-ICCVW2019}, MST++~\cite{MSTpp_cvpr2022}, ReSSS~\cite{ReSSS_TPAMI2024}, SSRT~\cite{SSRT_UNet_TGRS2024}, and SERT~\cite{SERT_CVPR2023}.

\subsubsection{Quantitative Evaluation}
Table~\ref{tab:denoising_comparison_clean} reports the full denoising grid over Gaussian noise and stripe-density levels. HAIR is best or second-best in most metrics; at $\sigma^2=1.0,\tilde{s}=0.1$, it improves PSNR over the second-best method (MST++) by more than 4 dB and reduces SAM by about 28.86\%. The gains at high interference indicate the benefit of adaptive destriping (Module A.II) and subspace denoising (Module A.III).

\begin{table*}[htbp]
    \centering
    \scriptsize
    \setlength{\tabcolsep}{2.2pt}
    \renewcommand{\arraystretch}{0.82}
    \caption{Quantitative comparison of different denoising methods across Gaussian noise and stripe-density levels.}
    \label{tab:denoising_comparison_clean}

    \begin{tabular}{cc c ccccc ccccc c}
        \toprule
        \multirow{2}{*}{Setup} & \multirow{2}{*}{Index} & \multirow{2}{*}{Noisy} & \multicolumn{5}{c}{Model-based Algorithms} & \multicolumn{5}{c}{Deep Learning Based Models} & \multirow{2}{*}{Ours}                                                                                                                                            \\
        \cmidrule(lr){4-8} \cmidrule(lr){9-13}
                               &                        &                        & SSTV                                       & LRTDTV                                         & E3DTV                 & NGMeet  & FHyDe               & DIP2d               & MST++               & ReSSS              & SSRT    & SERT                          \\
        \midrule
        \multirow{5}{*}{\makecell{$\sigma^2=0.1$                                                                                                                                                                                                                                                                                                  \\ $\tilde{s}=0.01$}}
                               & PSNR$\uparrow$         & 32.7398                & 39.5847                                    & 44.7983                                        & 44.9426               & 45.6458 & 46.9948             & \underline{52.8063} & 51.9315             & 45.7163            & 43.5747 & 38.2065 & \textbf{54.0590}    \\
                               & SSIM$\uparrow$         & 0.3375                 & 0.6143                                     & 0.9313                                         & 0.9690                & 0.9425  & 0.9341              & 0.9721              & \underline{0.9746}  & 0.8915             & 0.8804  & 0.5258  & \textbf{0.9892}     \\
                               & ERGAS$\downarrow$      & 2.6441                 & 1.2077                                     & 0.6512                                         & 0.7095                & 0.5958  & 0.5121              & \underline{0.2711}  & 0.2986              & 0.6049             & 0.7608  & 1.4227  & \textbf{0.2456}     \\
                               & RMSE$\downarrow$       & 0.2722                 & 0.1238                                     & 0.0680                                         & 0.0682                & 0.0627  & 0.0528              & \underline{0.0272}  & 0.0302              & 0.0615             & 0.0786  & 0.1456  & \textbf{0.0242}     \\
                               & SAM$\downarrow$        & 1.5044                 & 0.6830                                     & 0.3406                                         & 0.3491                & 0.1959  & 0.2573              & \underline{0.1276}  & 0.1299              & 0.3146             & 0.4074  & 0.7880  & \textbf{0.0940}     \\
        \midrule

        \multirow{5}{*}{\makecell{$\sigma^2=0.1$                                                                                                                                                                                                                                                                                                  \\ $\tilde{s}=0.05$}}
                               & PSNR$\uparrow$         & 31.4776                & 38.9341                                    & 44.4216                                        & 44.7179               & 45.3137 & 45.5484             & \underline{52.6892} & 51.3534             & 43.9890            & 41.6180 & 37.4639 & \textbf{54.0338}    \\
                               & SSIM$\uparrow$         & 0.2987                 & 0.5891                                     & 0.9133                                         & 0.9694                & 0.9402  & 0.9061              & \underline{0.9718}  & 0.9701              & 0.8630             & 0.8387  & 0.4868  & \textbf{0.9891}     \\
                               & ERGAS$\downarrow$      & 3.0679                 & 1.3052                                     & 0.6861                                         & 0.7209                & 0.6184  & 0.6049              & \underline{0.2780}  & 0.3173              & 0.7451             & 0.9580  & 1.5498  & \textbf{0.2461}     \\
                               & RMSE$\downarrow$       & 0.3148                 & 0.1334                                     & 0.0710                                         & 0.0700                & 0.0650  & 0.0625              & \underline{0.0277}  & 0.0322              & 0.0754             & 0.0984  & 0.1587  & \textbf{0.0243}     \\
                               & SAM$\downarrow$        & 1.7382                 & 0.7364                                     & 0.3564                                         & 0.3600                & 0.2098  & 0.2985              & \underline{0.1291}  & 0.1342              & 0.3429             & 0.5228  & 0.8589  & \textbf{0.0942}     \\
        \midrule

        \multirow{5}{*}{\makecell{$\sigma^2=0.1$                                                                                                                                                                                                                                                                                                  \\ $\tilde{s}=0.1$}}
                               & PSNR$\uparrow$         & 30.1137                & 38.0386                                    & 43.9862                                        & 44.5260               & 44.8587 & 44.1231             & \textbf{53.1534}    & 50.5455             & 43.2330            & 40.0738 & 36.5869 & \underline{52.4372} \\
                               & SSIM$\uparrow$         & 0.2485                 & 0.5536                                     & 0.8682                                         & 0.9673                & 0.9336  & 0.8573              & \underline{0.9733}  & 0.9625              & 0.8490             & 0.8072  & 0.4404  & \textbf{0.9814}     \\
                               & ERGAS$\downarrow$      & 3.5902                 & 1.4485                                     & 0.7188                                         & 0.7376                & 0.6508  & 0.7137              & \textbf{0.2607}     & 0.3459              & 0.8122             & 1.1491  & 1.7141  & \underline{0.2889}  \\
                               & RMSE$\downarrow$       & 0.3684                 & 0.1479                                     & 0.0747                                         & 0.0714                & 0.0683  & 0.0735              & \textbf{0.0260}     & 0.0352              & 0.0822             & 0.1175  & 0.1756  & \underline{0.0290}  \\
                               & SAM$\downarrow$        & 2.0334                 & 0.8164                                     & 0.3738                                         & 0.3655                & 0.2355  & 0.3632              & \underline{0.1233}  & 0.1423              & 0.3550             & 0.6234  & 0.9512  & \textbf{0.1014}     \\
        \midrule

        \multirow{5}{*}{\makecell{$\sigma^2=0.5$                                                                                                                                                                                                                                                                                                  \\ $\tilde{s}=0.01$}}
                               & PSNR$\uparrow$         & 26.9715                & 36.4156                                    & 43.2010                                        & 43.0166               & 45.4541 & 47.2024             & 42.7741             & \underline{49.3624} & 41.5084            & 40.7087 & 33.2990 & \textbf{51.9019}    \\
                               & SSIM$\uparrow$         & 0.1297                 & 0.4582                                     & 0.9096                                         & \underline{0.9571}    & 0.9436  & 0.9344              & 0.8508              & 0.9465              & 0.7740             & 0.7786  & 0.2546  & \textbf{0.9761}     \\
                               & ERGAS$\downarrow$      & 5.1550                 & 1.7408                                     & 0.7824                                         & 0.8698                & 0.6109  & 0.5007              & 0.8429              & \underline{0.3929}  & 0.9832             & 1.0610  & 2.5059  & \textbf{0.3039}     \\
                               & RMSE$\downarrow$       & 0.5288                 & 0.1783                                     & 0.0818                                         & 0.0846                & 0.0642  & 0.0515              & 0.0861              & \underline{0.0403}  & 0.0996             & 0.1090  & 0.2563  & \textbf{0.0308}     \\
                               & SAM$\downarrow$        & 2.9310                 & 0.9830                                     & 0.4175                                         & 0.4224                & 0.1957  & 0.2467              & 0.4580              & \underline{0.1565}  & 0.5224             & 0.5753  & 1.4041  & \textbf{0.1165}     \\
        \midrule

        \multirow{5}{*}{\makecell{$\sigma^2=0.5$                                                                                                                                                                                                                                                                                                  \\ $\tilde{s}=0.05$}}
                               & PSNR$\uparrow$         & 26.5609                & 36.2381                                    & 42.5079                                        & 42.7091               & 45.1720 & 45.7891             & 43.2288             & \underline{49.1939} & 40.8889            & 39.7744 & 33.2134 & \textbf{51.7969}    \\
                               & SSIM$\uparrow$         & 0.1188                 & 0.4444                                     & 0.8920                                         & \underline{0.9555}    & 0.9431  & 0.9051              & 0.8559              & 0.9428              & 0.7579             & 0.7546  & 0.2518  & \textbf{0.9758}     \\
                               & ERGAS$\downarrow$      & 5.4216                 & 1.8259                                     & 0.8682                                         & 0.8807                & 0.6300  & 0.5902              & 0.8142              & \underline{0.4005}  & 1.0595             & 1.1865  & 2.5338  & \textbf{0.3073}     \\
                               & RMSE$\downarrow$       & 0.5545                 & 0.1866                                     & 0.0885                                         & 0.0876                & 0.0662  & 0.0607              & 0.0820              & \underline{0.0410}  & 0.1069             & 0.1215  & 0.2590  & \textbf{0.0312}     \\
                               & SAM$\downarrow$        & 3.0721                 & 1.0290                                     & 0.4536                                         & 0.4338                & 0.2083  & 0.2891              & 0.4340              & \underline{0.1565}  & 0.5353             & 0.6453  & 1.4193  & \textbf{0.1180}     \\
        \midrule

        \multirow{5}{*}{\makecell{$\sigma^2=0.5$                                                                                                                                                                                                                                                                                                  \\ $\tilde{s}=0.1$}}
                               & PSNR$\uparrow$         & 25.9727                & 35.4320                                    & 42.6654                                        & 42.4968               & 45.1567 & 44.1921             & 42.7499             & \underline{47.7716} & 40.5923            & 38.6110 & 32.8830 & \textbf{51.3960}    \\
                               & SSIM$\uparrow$         & 0.1000                 & 0.4218                                     & 0.8408                                         & \underline{0.9547}    & 0.9430  & 0.8578              & 0.8444              & 0.9293              & 0.7550             & 0.7264  & 0.2393  & \textbf{0.9749}     \\
                               & ERGAS$\downarrow$      & 5.8008                 & 1.9553                                     & 0.8380                                         & 0.9178                & 0.6356  & 0.7096              & 0.8436              & \underline{0.4721}  & 1.0972             & 1.3580  & 2.6320  & \textbf{0.3205}     \\
                               & RMSE$\downarrow$       & 0.5933                 & 0.1997                                     & 0.0869                                         & 0.0897                & 0.0664  & 0.0730              & 0.0861              & \underline{0.0486}  & 0.1107             & 0.1388  & 0.2691  & \textbf{0.0326}     \\
                               & SAM$\downarrow$        & 3.2859                 & 1.1017                                     & 0.4411                                         & 0.4466                & 0.2266  & 0.3561              & 0.4562              & \underline{0.1780}  & 0.5396             & 0.7417  & 1.4740  & \textbf{0.1257}     \\
        \midrule

        \multirow{5}{*}{\makecell{$\sigma^2=1.0$                                                                                                                                                                                                                                                                                                  \\ $\tilde{s}=0.01$}}
                               & PSNR$\uparrow$         & 24.1413                & 35.0929                                    & 41.6157                                        & 42.0517               & 45.3088 & \underline{47.4565} & 38.8767             & 46.9902             & 39.4439            & 37.5132 & 31.5225 & \textbf{51.3127}    \\
                               & SSIM$\uparrow$         & 0.0788                 & 0.4008                                     & 0.8613                                         & \underline{0.9526}    & 0.9430  & 0.9363              & 0.7819              & 0.9180              & 0.7069             & 0.6860  & 0.1834  & \textbf{0.9724}     \\
                               & ERGAS$\downarrow$      & 7.1450                 & 2.0275                                     & 0.9559                                         & 0.9718                & 0.6207  & \underline{0.4870}  & 1.3138              & 0.5121              & 1.2513             & 1.5415  & 3.0795  & \textbf{0.3224}     \\
                               & RMSE$\downarrow$       & 0.7326                 & 0.2076                                     & 0.0980                                         & 0.0945                & 0.0650  & \underline{0.0501}  & 0.1347              & 0.0530              & 0.1261             & 0.1574  & 0.3151  & \textbf{0.0328}     \\
                               & SAM$\downarrow$        & 4.0577                 & 1.1444                                     & 0.5034                                         & 0.4702                & 0.2011  & 0.2363              & 0.7222              & \underline{0.1850}  & 0.6577             & 0.8446  & 1.7282  & \textbf{0.1261}     \\
        \midrule

        \multirow{5}{*}{\makecell{$\sigma^2=1.0$                                                                                                                                                                                                                                                                                                  \\ $\tilde{s}=0.05$}}
                               & PSNR$\uparrow$         & 23.8964                & 34.7828                                    & 41.4754                                        & 41.7788               & 44.9932 & 46.2746             & 39.2177             & \underline{47.1948} & 39.8763            & 37.1702 & 31.5733 & \textbf{51.2096}    \\
                               & SSIM$\uparrow$         & 0.0736                 & 0.3908                                     & 0.8535                                         & \underline{0.9505}    & 0.9420  & 0.9138              & 0.7761              & 0.9148              & 0.7243             & 0.6770  & 0.1856  & \textbf{0.9722}     \\
                               & ERGAS$\downarrow$      & 7.3734                 & 2.1062                                     & 0.9925                                         & 0.9723                & 0.6439  & 0.5598              & 1.3423              & \underline{0.5003}  & 1.2042             & 1.6040  & 3.0616  & \textbf{0.3260}     \\
                               & RMSE$\downarrow$       & 0.7535                 & 0.2152                                     & 0.0996                                         & 0.0974                & 0.0675  & 0.0575              & 0.1348              & \underline{0.0516}  & 0.1210             & 0.1637  & 0.3134  & \textbf{0.0332}     \\
                               & SAM$\downarrow$        & 4.1733                 & 1.1864                                     & 0.5093                                         & 0.4795                & 0.2132  & 0.2714              & 0.7287              & \underline{0.1781}  & 0.6268             & 0.8799  & 1.7173  & \textbf{0.1278}     \\
        \midrule

        \multirow{5}{*}{\makecell{$\sigma^2=1.0$                                                                                                                                                                                                                                                                                                  \\ $\tilde{s}=0.1$}}
                               & PSNR$\uparrow$         & 23.4978                & 34.2974                                    & 41.0582                                        & 41.5646               & 44.6525 & 44.4068             & 39.2021             & \underline{46.6157} & 39.3400            & 37.3486 & 31.5275 & \textbf{50.7045}    \\
                               & SSIM$\uparrow$         & 0.0617                 & 0.3731                                     & 0.8215                                         & \underline{0.9496}    & 0.9416  & 0.8668              & 0.7714              & 0.9081              & 0.7181             & 0.6699  & 0.1838  & \textbf{0.9710}     \\
                               & ERGAS$\downarrow$      & 7.7198                 & 2.2279                                     & 1.0061                                         & 1.0210                & 0.6660  & 0.6932              & 1.2910              & 1.3028              & \underline{0.5329} & 1.5715  & 3.0775  & \textbf{0.3442}     \\
                               & RMSE$\downarrow$       & 0.7889                 & 0.2275                                     & 0.1045                                         & 0.0999                & 0.0699  & 0.0712              & 0.1293              & 0.1325              & \underline{0.0552} & 0.1604  & 0.3149  & \textbf{0.0351}     \\
                               & SAM$\downarrow$        & 4.3680                 & 1.2550                                     & 0.5334                                         & 0.4941                & 0.2328  & 0.3436              & 0.6437              & 0.7116              & \underline{0.1954} & 0.8594  & 1.7271  & \textbf{0.1390}     \\
        \bottomrule
    \end{tabular}
    \\[2pt]
    \parbox{\linewidth}{\centering \footnotesize
        The best and second-best results are highlighted in \textbf{bold} and \underline{underlined}. ``$\uparrow$'' (resp. ``$\downarrow$'') means the larger (resp. smaller), the better. PSNR is in dB.
    }
\end{table*}

\begin{figure}[htbp]
    \centering
    \includegraphics[width=1\linewidth]{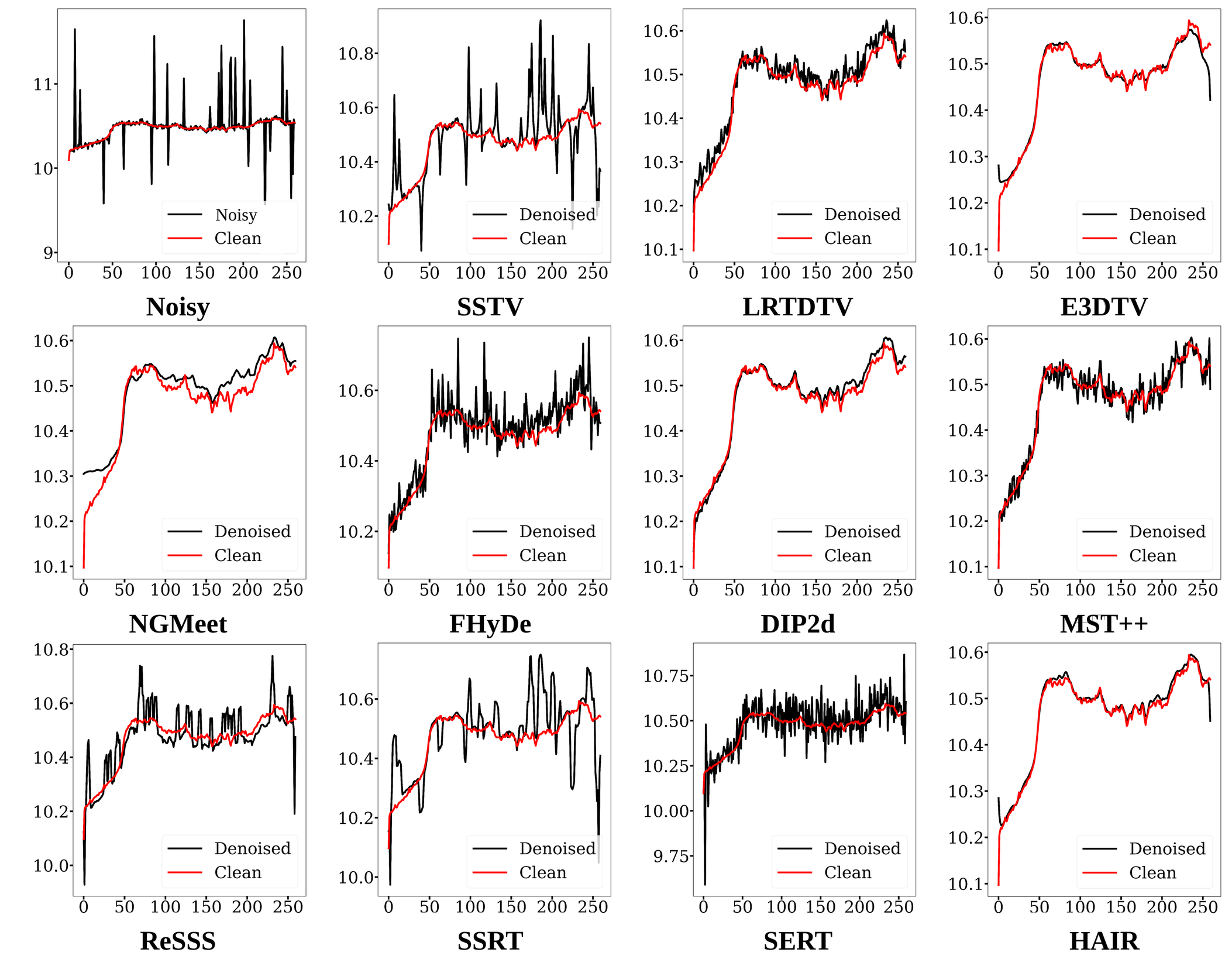}
    \caption{Horizontal mean curves for central band 128 reconstructed by methods in experiment ($\sigma=1.0, \tilde{s}=0.1$).  y-axis: mean radiance; x-axis: row index.}
    \label{fig:horizontal_mean_denoising}
\end{figure}

\begin{figure}[htbp]
    \centering
    \includegraphics[width=1\linewidth]{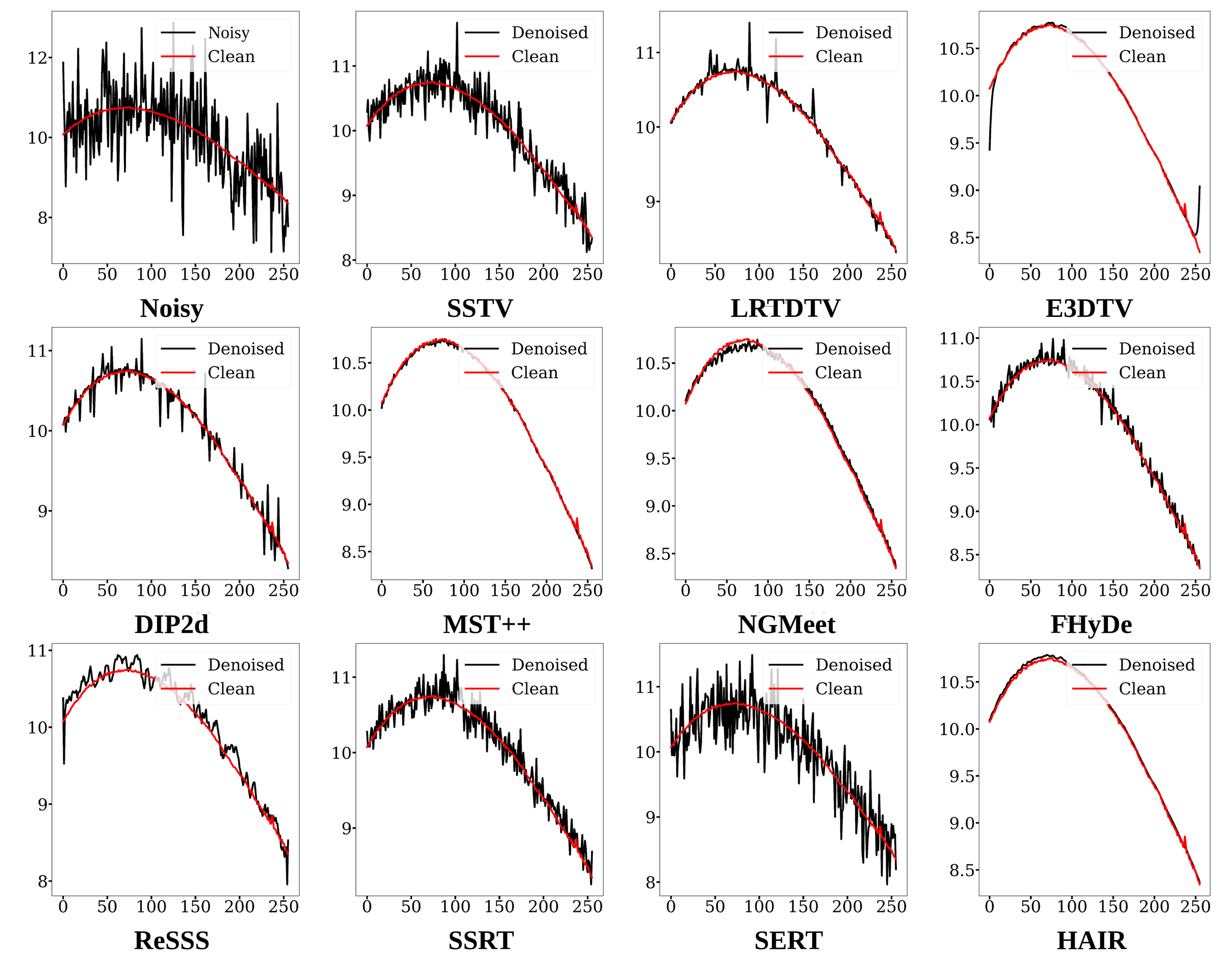}
    \caption{Spectral signatures of central pixel $(130, 750)$ restored by methods in experiment ($\sigma=1.0, \tilde{s}=0.1$).  y-axis: radiance; x-axis: band index.}
    \label{fig:spectral_curve_denoising}
\end{figure}

\subsubsection{Qualitative Visual and Physical Retrieval Analysis}
\begin{figure*}[htbp]
    \centering
    \includegraphics[width=1\linewidth, height=0.96\textheight, keepaspectratio]{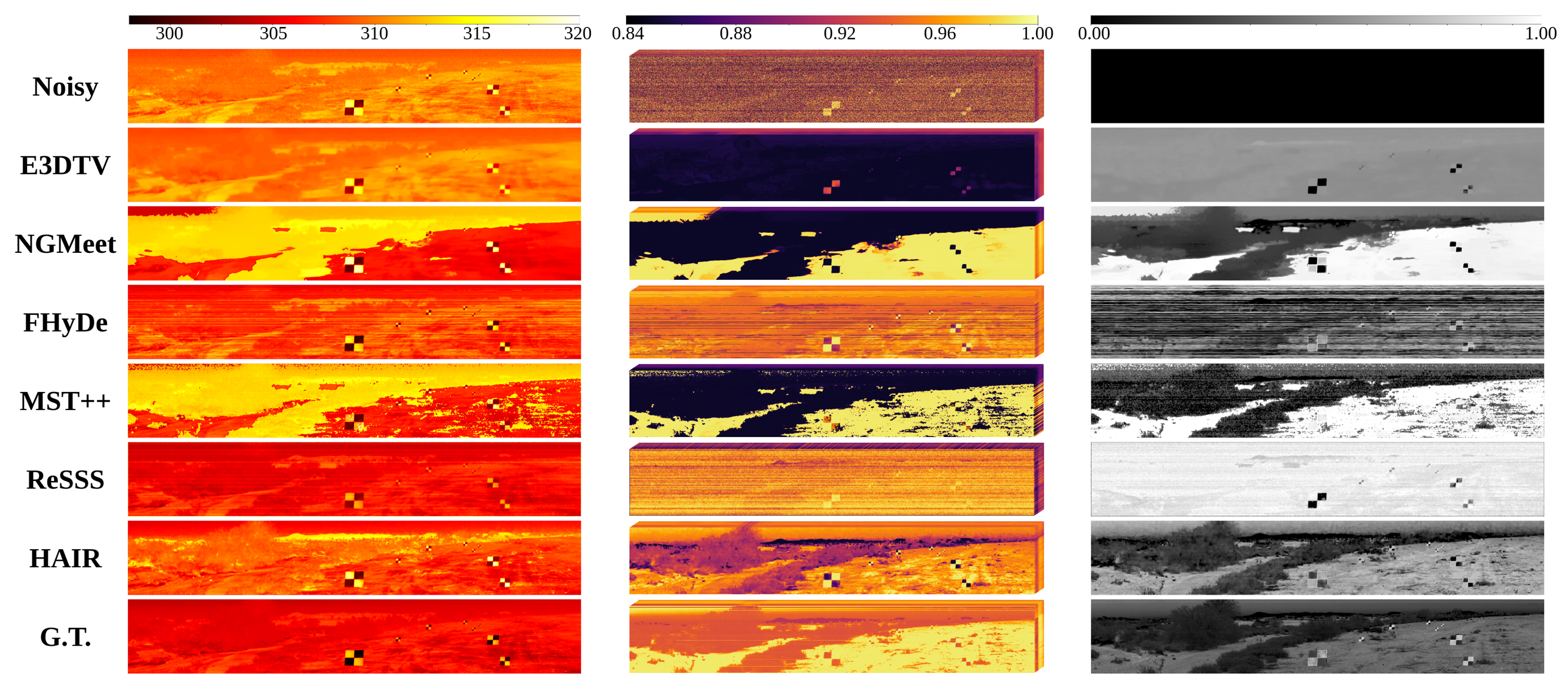}
    \caption{HADAR retrievals in the denoising experiment ($\sigma=1.0, \tilde{s} = 0.1$). Columns 1--3 show temperature ($T$, in K), emissivity ($e$), and normalized texture ($X$), respectively.}
    \label{fig:simulated_denoising_etc_band128}
\end{figure*}

\begin{figure*}[htbp]
    \centering
    \includegraphics[width=1\linewidth, height=0.96\textheight, keepaspectratio]{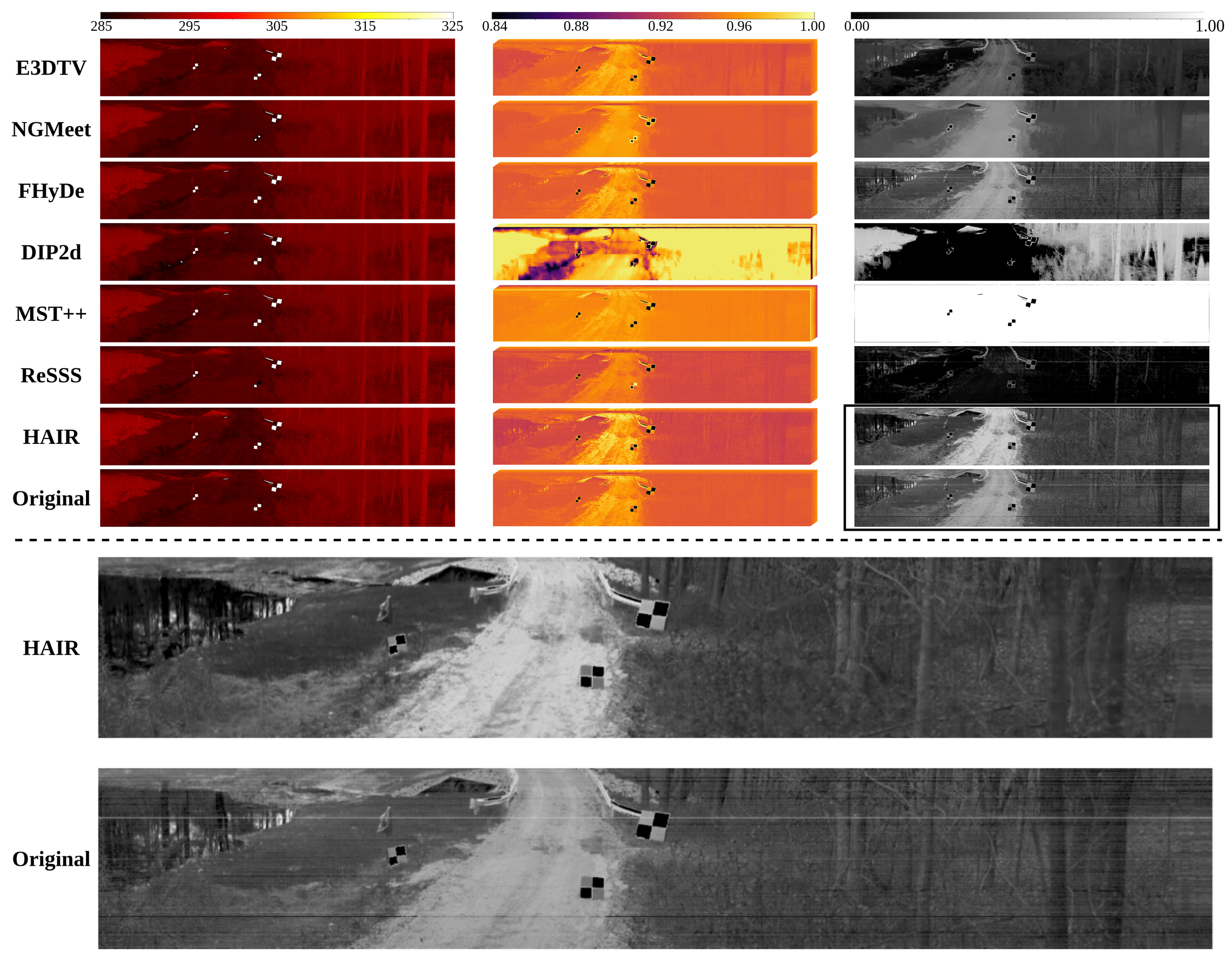}
    \caption{HADAR retrievals on real-world pushbroom imagery. Columns 1--3 show temperature ($T$, in K), emissivity ($e$), and normalized texture ($X$), respectively. The bottom panels provide zoomed-in views of the normalized texture $X$ for HAIR and the original HSI.}
    \label{fig:real_world_denoising}
\end{figure*}

\begin{figure*}
    \centering
    \includegraphics[width=1\linewidth]{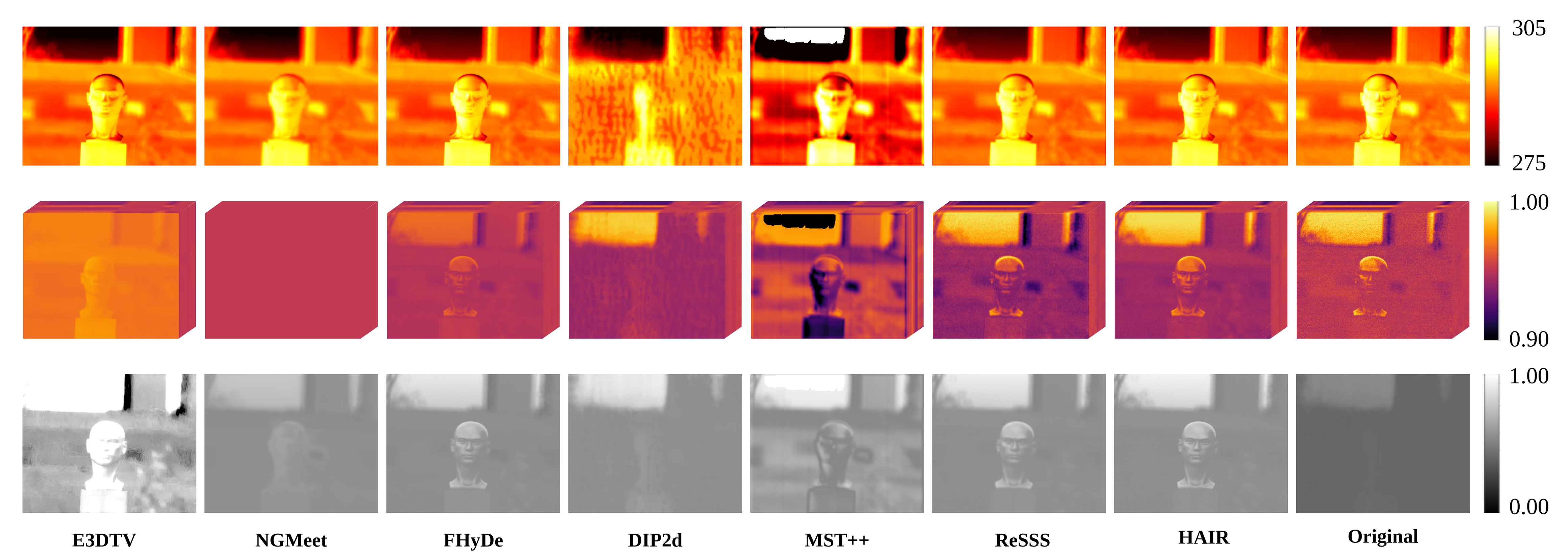}
    \caption{HADAR retrievals on real-world FTIR measurements. Rows 1--3 show temperature ($T$, in K), emissivity ($e$), and normalized texture ($X$), respectively.}
    \label{fig:denoised_whu_etc}
\end{figure*}

Figures~\ref{fig:simulated_denoising_etc_band128}, \ref{fig:real_world_denoising}, and \ref{fig:denoised_whu_etc} show simulated IH-10, real pushbroom, and real FTIR retrievals. Although smoothing-based and deep spectral-spatial methods can be competitive in Table~\ref{tab:denoising_comparison_clean}, their restored spectra often suppress high-frequency atmospheric structures or distort band transitions, as shown in Figs.~\ref{fig:horizontal_mean_denoising} and \ref{fig:spectral_curve_denoising}. These errors propagate to HADAR inversion, whereas HAIR better preserves the spatial-spectral profiles and yields more plausible $e$, $T$, and $X$ maps across the three visual settings.

\subsection{TIR-HSI Inpainting Experiments}
We compare HAIR with model-based methods E3DTV~\cite{E3DTV_TIP2020}, NGMeet~\cite{NGMeet_hwwei_tpami_2022}, and FHyIn~\cite{ZhuangFastHyperspectralImage2018}, and deep methods HLRTF~\cite{HLRTF_CVPR2022}, DIP2d/DIP3d~\cite{DIP2D-DIP3D-ICCVW2019}, MST++~\cite{MSTpp_cvpr2022}, SSRT~\cite{SSRT_UNet_TGRS2024}, SERT~\cite{SERT_CVPR2023}, and ReSSS~\cite{ReSSS_TPAMI2024} in catastrophic band-wise corruption.

\subsubsection{Quantitative Evaluation}
\begin{figure}[htbp]
    \centering
    \includegraphics[width=1\linewidth]{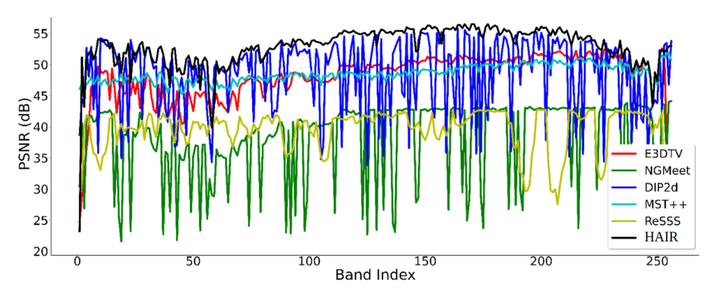}
    \caption{Band-wise PSNR comparison for the inpainting task under $\tilde{b}=0.2$.}
    \label{fig:inpaint02_psnr}
\end{figure}

As reported in Table~\ref{tab:inpainting_comparison_clean}, HAIR outperforms competing methods across all corruption levels. The band-wise PSNR curves in Fig.~\ref{fig:inpaint02_psnr} further show stable missing-band recovery without performance collapse.

\begin{table*}[htbp]
    \centering
    \scriptsize
    \setlength{\tabcolsep}{2.2pt}
    \renewcommand{\arraystretch}{0.82}
    \caption{Quantitative comparison of different inpainting methods across missing levels.}
    \label{tab:inpainting_comparison_clean}
    \begin{tabular}{cc c ccccc ccccc c}
        \toprule
        \multirow{2}{*}{Setup} & \multirow{2}{*}{Index} & \multirow{2}{*}{Noisy} & \multicolumn{3}{c}{Model-based Algorithms} & \multicolumn{7}{c}{Deep Learning Based Models} & \multirow{2}{*}{Ours}                                                                                                                  \\
        \cmidrule(lr){4-6} \cmidrule(lr){7-13}
                               &                        &                        & E3DTV                                      & NGMeet                                         & FHyIn                 & HLRTF   & DIP2d               & DIP3d   & MST++               & SSRT    & SERT    & ReSSS                      \\
        \midrule
        \multirow{5}{*}{\makecell{$\tilde{b}=0.2$}}
                               & PSNR$\uparrow$         & 19.5281                & 42.3335                                    & 32.9267                                        & 28.0998               & 28.1605 & 47.2119             & 36.3671 & \underline{47.6966} & 33.0055 & 31.7435 & 36.6406 & \textbf{52.2184} \\
                               & SSIM$\uparrow$         & 0.1150                 & \underline{0.9566}                         & 0.7528                                         & 0.6372                & 0.1834  & 0.9267              & 0.9317  & 0.9246              & 0.7074  & 0.2079  & 0.7332  & \textbf{0.9847}  \\
                               & ERGAS$\downarrow$      & 12.6727                & 0.9144                                     & 2.8614                                         & 4.4575                & 4.5276  & 0.5184              & 2.4741  & \underline{0.4737}  & 2.6242  & 3.0273  & 1.7865  & \textbf{0.2935}  \\
                               & RMSE$\downarrow$       & 1.3054                 & 0.0915                                     & 0.3014                                         & 0.4650                & 0.4612  & 0.0526              & 0.2636  & \underline{0.0488}  & 0.2656  & 0.3103  & 0.1751  & \textbf{0.0296}  \\
                               & SAM$\downarrow$        & 7.1176                 & 0.4616                                     & 1.4507                                         & 2.3581                & 2.5552  & 0.2385              & 0.8105  & \underline{0.1800}  & 1.4332  & 1.6846  & 0.9223  & \textbf{0.1166}  \\
        \midrule

        \multirow{5}{*}{\makecell{$\tilde{b}=0.3$}}
                               & PSNR$\uparrow$         & 17.6204                & 42.2670                                    & 29.8769                                        & 27.1996               & 28.1547 & \underline{46.4340} & 32.1854 & 45.7544             & 31.7052 & 30.0550 & 36.2579 & \textbf{52.2215} \\
                               & SSIM$\uparrow$         & 0.1108                 & \underline{0.9550}                         & 0.6475                                         & 0.5379                & 0.1829  & 0.9254              & 0.9231  & 0.9203              & 0.6608  & 0.1584  & 0.7525  & \textbf{0.9845}  \\
                               & ERGAS$\downarrow$      & 15.1171                & 0.9368                                     & 3.6382                                         & 4.9719                & 4.5303  & \underline{0.5910}  & 4.2260  & 0.5949              & 3.0411  & 3.6712  & 1.9196  & \textbf{0.2942}  \\
                               & RMSE$\downarrow$       & 1.5532                 & 0.0921                                     & 0.3806                                         & 0.5157                & 0.4615  & 0.0608              & 0.4076  & \underline{0.0615}  & 0.3087  & 0.3766  & 0.1865  & \textbf{0.0296}  \\
                               & SAM$\downarrow$        & 8.4639                 & 0.4600                                     & 1.8573                                         & 2.5999                & 2.5565  & \underline{0.2466}  & 1.7261  & 0.9203              & 1.6740  & 2.0425  & 0.9566  & \textbf{0.1146}  \\
        \midrule

        \multirow{5}{*}{\makecell{$\tilde{b}=0.5$}}
                               & PSNR$\uparrow$         & 15.4292                & 41.6048                                    & 27.4976                                        & 26.0497               & 28.8741 & \underline{46.6349} & 34.4997 & 43.2806             & 29.8485 & 29.5427 & 35.4133 & \textbf{49.6242} \\
                               & SSIM$\uparrow$         & 0.0717                 & \underline{0.9518}                         & 0.4223                                         & 0.3389                & 0.1956  & 0.9187              & 0.9267  & 0.8946              & 0.5832  & 0.1473  & 0.7405  & \textbf{0.9837}  \\
                               & ERGAS$\downarrow$      & 19.3618                & 1.0231                                     & 4.7713                                         & 5.6548                & 4.1703  & \underline{0.5449}  & 2.8455  & 0.7808              & 3.7842  & 3.9035  & 2.1822  & \textbf{0.4278}  \\
                               & RMSE$\downarrow$       & 1.9991                 & 0.0996                                     & 0.4995                                         & 0.5887                & 0.4248  & \underline{0.0556}  & 0.2952  & 0.0818              & 0.3825  & 0.4006  & 0.2072  & \textbf{0.0443}  \\
                               & SAM$\downarrow$        & 10.8849                & 0.4983                                     & 2.5158                                         & 2.9712                & 2.3513  & 0.2619              & 1.0182  & \underline{0.2438}  & 2.0747  & 2.1581  & 1.1116  & \textbf{0.1969}  \\

        \bottomrule
    \end{tabular}
    \\[2pt]
    \parbox{\linewidth}{\centering \footnotesize
        The best and second-best results are highlighted in \textbf{bold} and \underline{underlined}, respectively. ``$\uparrow$'' (resp. ``$\downarrow$'') means the larger (resp. smaller), the better.
    }
\end{table*}

\subsubsection{Spectral Reconstruction Fidelity and Physical Retrieval Consistency}
\begin{figure*}[htbp]
    \centering
    \includegraphics[width=1\linewidth, height=0.96\textheight, keepaspectratio]{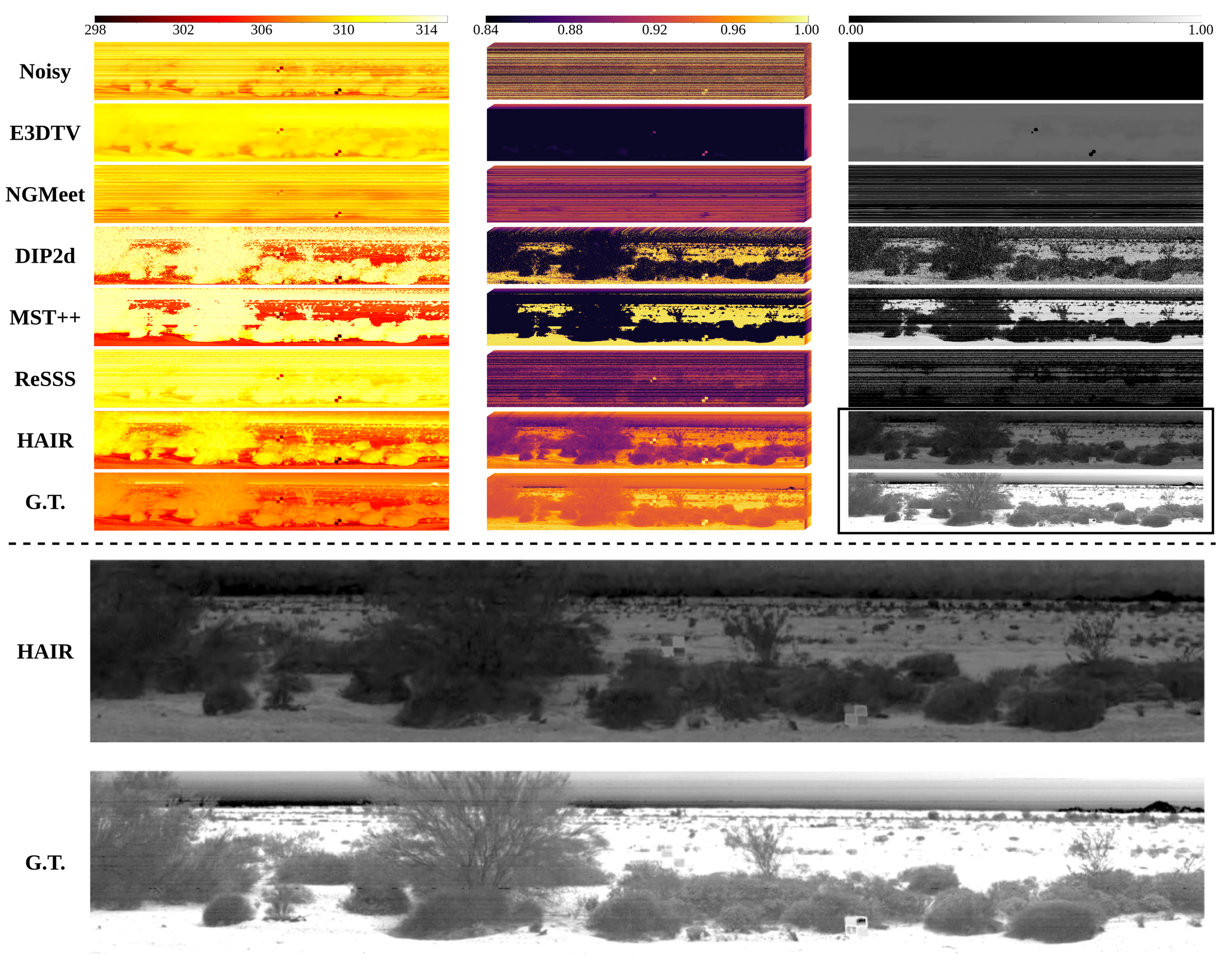}
    \caption{HADAR retrievals in the inpainting experiment ($ \tilde{b} = 0.5$). Columns 1--3 show temperature ($T$, in K), emissivity ($e$), and normalized texture ($X$), respectively. The bottom panels provide zoomed-in views of the normalized texture $X$ for HAIR and the ground truth HSI.}
    \label{fig:inpainting_etc}
\end{figure*}

Figure~\ref{fig:inpainting_etc} shows that severe missing bands also affect physical retrieval: several baselines produce plausible radiance but implausible TeX combinations, such as high temperature with underestimated emissivity. This conflicts with material priors from the NASA spectral library~\cite{NASA_ECOSTRESS_SpectralLib_2019}, where grass and sand emissivity in LWIR is generally close to 0.95. HAIR better restores missing spectra while maintaining physically credible emissivity, temperature, and texture maps.

\subsection{TIR-HSI Spectral Calibration Experiments}
\begin{figure}
    \centering
    \includegraphics[width=1\linewidth]{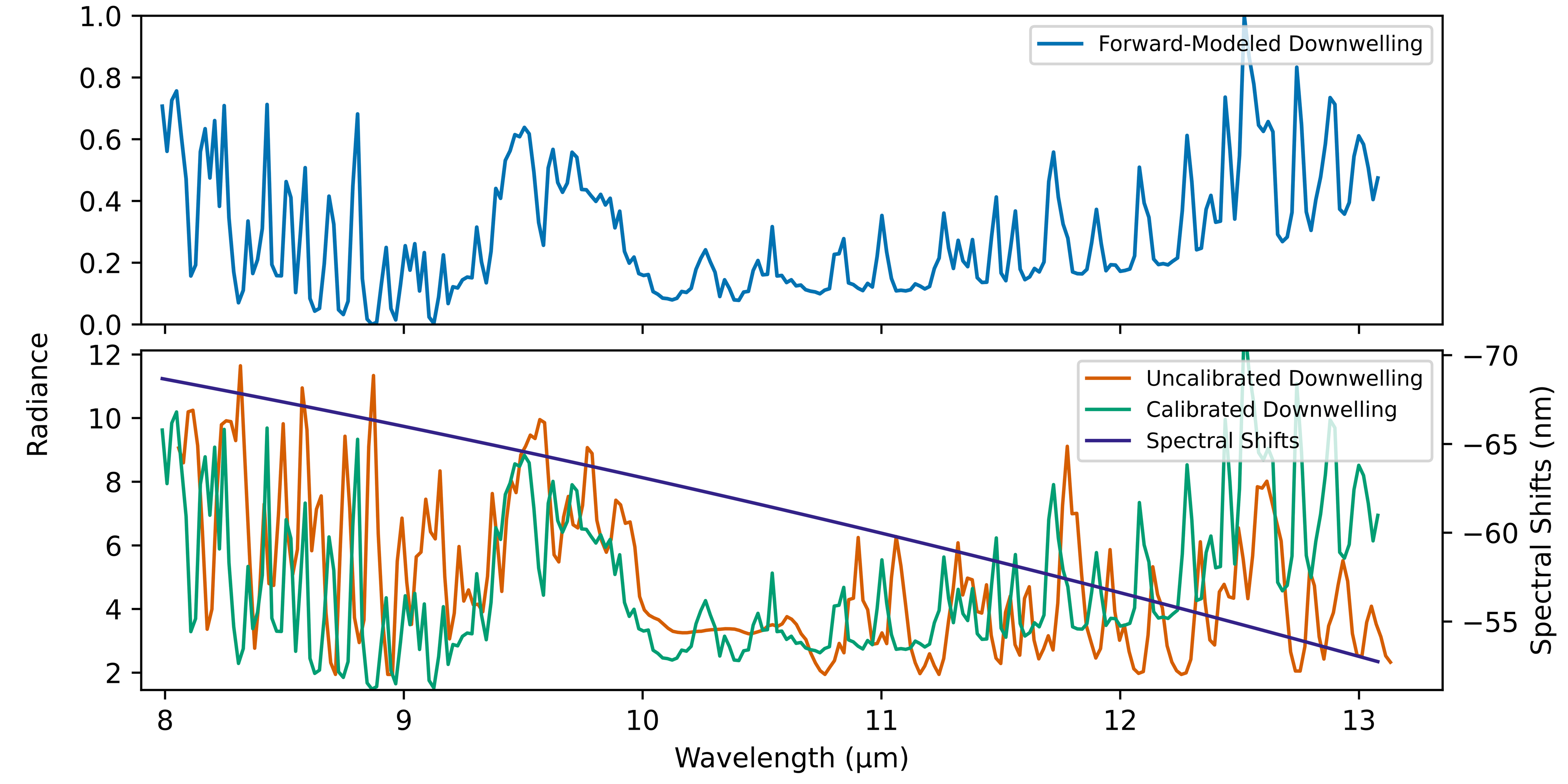}
    \caption{Spectral calibration process. The uncalibrated observed sky signal $\mathbf{s}$ is aligned against a forward-modeled reference $\mathbf{s}_{\mathrm{r}}$ (Eq.~\ref{eq:spectral_correction}) to compute the spectral shift map $\Delta \lambda_k$ (Eq.~\ref{eq:shift_map}) and the calibrated sky signal $\hat{\mathbf{s}}(k)$ (Eq.~\ref{eq:calibrated_sky}).}
    \label{fig:spectral_correction_comparison}
\end{figure}

\begin{figure}
    \centering
    \includegraphics[width=1\linewidth]{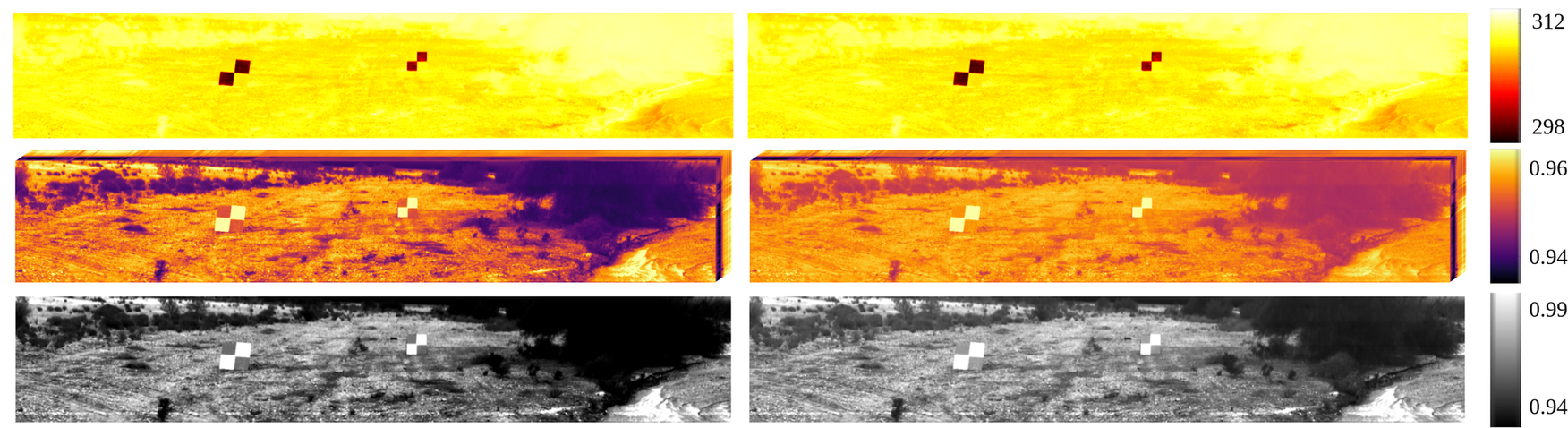}
    \caption{HADAR retrievals comparison on uncalibrated and calibrated HSI. Left: Uncalibrated. Right: Calibrated. Rows 1--3 show temperature ($T$ in K), emissivity ($e$), and normalized texture ($X$), respectively.}
    \label{fig:etx_calibration}
\end{figure}

We verify spectral calibration on IH-10 through controlled wavelength-shift simulation. Table~\ref{tab:calibration_necessity} shows that neglecting band correction introduces TeX and HSI errors, including about 0.2 K average temperature deviation and 0.01 average emissivity error. Although numerically small, such deviations matter for TIR scenes with similar emissivity and close temperature ranges. Figures~\ref{fig:spectral_correction_comparison} and \ref{fig:etx_calibration} further show that calibration aligns the sky signature, improves emissivity and texture plausibility, and preserves stable temperature retrieval.

\begin{table}[htbp]
    \scriptsize
    \setlength{\tabcolsep}{2.5pt}
    \renewcommand{\arraystretch}{0.95}
    \caption{Effect of spectral calibration on HADAR inversion.}
    \label{tab:calibration_necessity}
    \centering
    \begin{tabular}{c cc cc cc cc}
        \toprule
        \multirow{2}{*}{Index}
                  & \multicolumn{2}{c}{$T$}
                  & \multicolumn{2}{c}{$e$}
                  & \multicolumn{2}{c}{$X$}
                  & \multicolumn{2}{c}{HSI}                                                                \\
        \cmidrule(lr){2-3} \cmidrule(lr){4-5} \cmidrule(lr){6-7} \cmidrule(lr){8-9}
                  & MSE                     & MAE
                  & MSE                     & MAE
                  & MSE                     & MAE
                  & MSE                     & MAE                                                          \\
        \midrule
        Deviation & 0.2063                  & 0.1690 & 0.0094 & 0.0068 & 0.0075 & 0.0067 & 0.0847 & 0.0582 \\
        \bottomrule
    \end{tabular}
\end{table}

\subsection{TIR-HSI Spectral Super-Resolution Experiments}
We evaluate SSR under severe spectral undersampling and compare HAIR with five HSI SSR baselines: LTRN~\cite{LTRN_TNN2025}, MST++~\cite{MSTpp_cvpr2022}, DRCR~\cite{DRCRNet_cvprw2022}, HSRNet~\cite{HSRNet_TNN2022}, and AWAN~\cite{AWAN_CVPRW2020}.

\subsubsection{Quantitative Comparison}
Table~\ref{tab:ssr_experiment} reports performance at $\times2$, $\times4$, and $\times8$ scales. HAIR consistently leads in PSNR, ERGAS, RMSE, and SAM; relative to the second-best model, its ERGAS and SAM improve by 23.59\% and 27.06\%, respectively, indicating stronger spectral fidelity.

\begin{table}[htbp]
    \scriptsize
    \setlength{\tabcolsep}{2.5pt}
    \renewcommand{\arraystretch}{0.95}
    \caption{Quantitative comparison of spectral super-resolution methods across scale levels.}
    \label{tab:ssr_experiment}
    \centering
    \begin{tabular}{c c c c c c c}
        \toprule
        Method  & Scale      & PSNR$\uparrow$      & SSIM$\uparrow$     & ERGAS$\downarrow$  & RMSE$\downarrow$   & SAM$\downarrow$    \\

        \midrule
        Bicubic & $\times 2$ & 32.2638             & 0.2406             & 2.8129             & 0.2876             & 1.5832             \\
        LTRN    & $\times 2$ & 43.2406             & 0.9197             & 0.8188             & 0.0825             & 0.4173             \\
        MST++   & $\times 2$ & 44.8247             & 0.9420             & 0.6747             & 0.0680             & 0.3387             \\
        DRCR    & $\times 2$ & 43.5202             & \underline{0.9578} & 0.7833             & 0.0807             & 0.3743             \\
        HSRNet  & $\times 2$ & 43.5579             & 0.9516             & 0.7508             & 0.0792             & \underline{0.2661} \\
        AWAN    & $\times 2$ & \underline{46.6841} & \textbf{0.9720}    & \underline{0.5411} & \underline{0.0559} & 0.2692             \\
        Ours    & $\times 2$ & \textbf{48.6776}    & 0.9454             & \textbf{0.4338}    & \textbf{0.0441}    & \textbf{0.1870}    \\

        \midrule
        Bicubic & $\times 4$ & 32.5587             & 0.2529             & 2.7173             & 0.2780             & 1.5248             \\
        LTRN    & $\times 4$ & 44.0036             & 0.9300             & 0.7558             & 0.0766             & 0.3816             \\
        MST++   & $\times 4$ & 45.7288             & 0.9498             & 0.6048             & 0.0614             & 0.2993             \\
        DRCR    & $\times 4$ & 43.4035             & 0.9561             & 0.7939             & 0.0818             & 0.3743             \\
        HSRNet  & $\times 4$ & 44.7288             & 0.9556             & 0.6537             & 0.0689             & \underline{0.2381} \\
        AWAN    & $\times 4$ & \underline{46.6550} & \underline{0.9722} & \underline{0.5423} & \underline{0.0559} & 0.2726             \\
        Ours    & $\times 4$ & \textbf{50.2725}    & \textbf{0.9737}    & \textbf{0.3661}    & \textbf{0.0373}    & \textbf{0.1478}    \\
        \midrule

        Bicubic & $\times 8$ & 32.6465             & 0.2575             & 2.6867             & 0.2752             & 1.4877             \\
        LTRN    & $\times 8$ & 43.8886             & 0.9331             & 0.7690             & 0.0793             & 0.3895             \\
        MST++   & $\times 8$ & 45.8150             & 0.9599             & 0.6051             & 0.0611             & 0.3065             \\
        DRCR    & $\times 8$ & 43.2811             & 0.9541             & 0.8036             & 0.0829             & 0.3724             \\
        HSRNet  & $\times 8$ & \underline{47.0962} & 0.9625             & \underline{0.5086} & \underline{0.0528} & \underline{0.1950} \\
        AWAN    & $\times 8$ & 46.2635             & \textbf{0.9693}    & 0.5743             & 0.0591             & 0.2819             \\
        Ours    & $\times 8$ & \textbf{49.0819}    & \underline{0.9689} & \textbf{0.4165}    & \textbf{0.0426}    & \textbf{0.1752}    \\

        \bottomrule
    \end{tabular}
    \\[2pt]
    \parbox{\linewidth}{\centering \footnotesize
        The best and second-best results are highlighted in \textbf{bold} and \underline{underlined}. ``$\uparrow$'' (resp. ``$\downarrow$'') means the larger (resp. smaller), the better. PSNR is in dB.
    }
\end{table}

\subsubsection{Spectral Reconstruction Fidelity and Physical Retrieval Consistency}

\begin{figure}[htbp]
    \centering
    \includegraphics[width=1\linewidth]{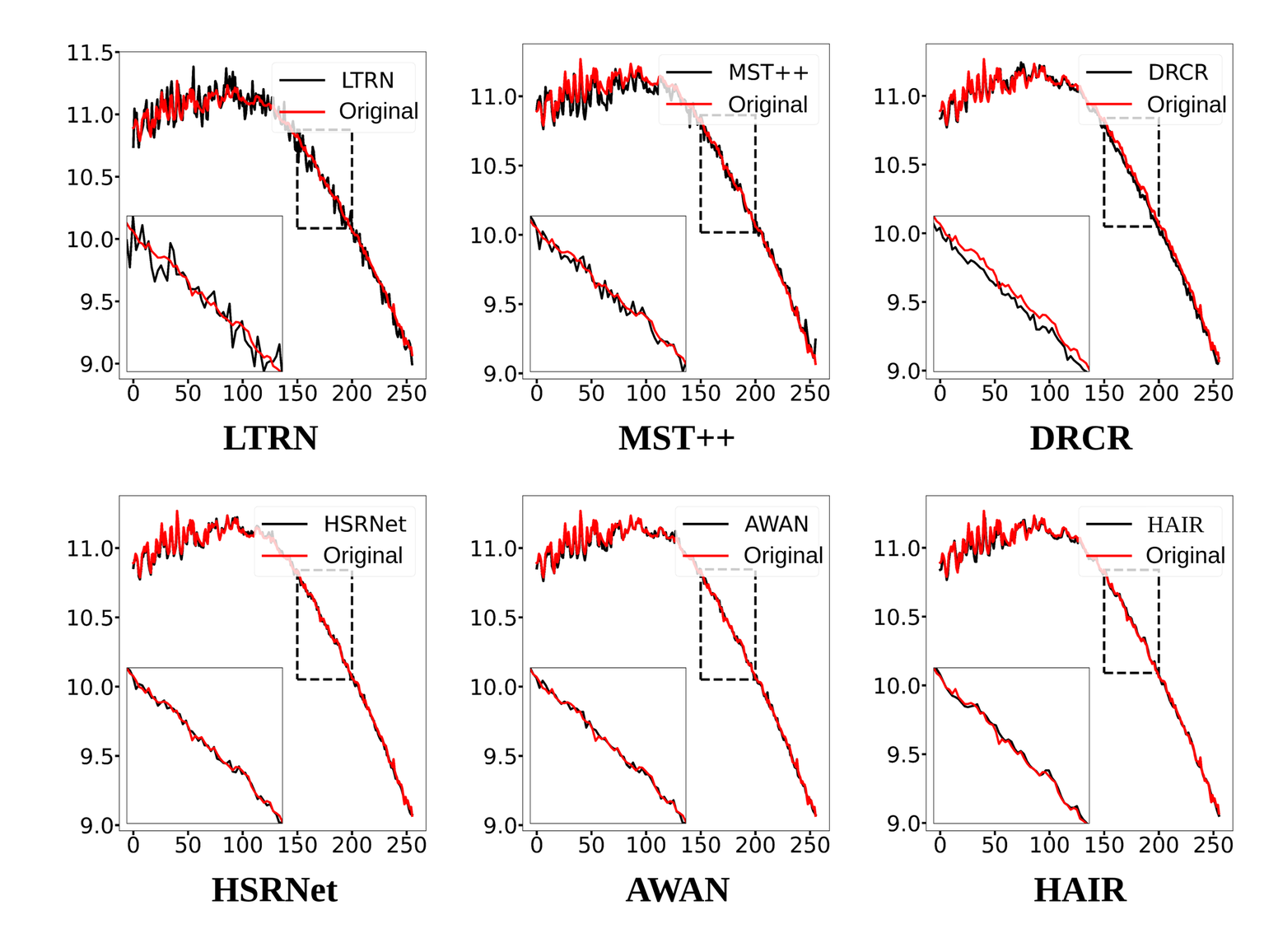}
    \caption{Comparison of spectral signatures for the central pixel $(130, 750)$ under the $8\times$ downsampling scale. Zoomed-in boxes are provided for detailed inspection of the reconstruction errors. y-axis: radiance; x-axis: band index.}
    \label{fig:ssr_reflective}
\end{figure}

\begin{figure*}[htbp]
    \centering
    \includegraphics[width=1\linewidth]{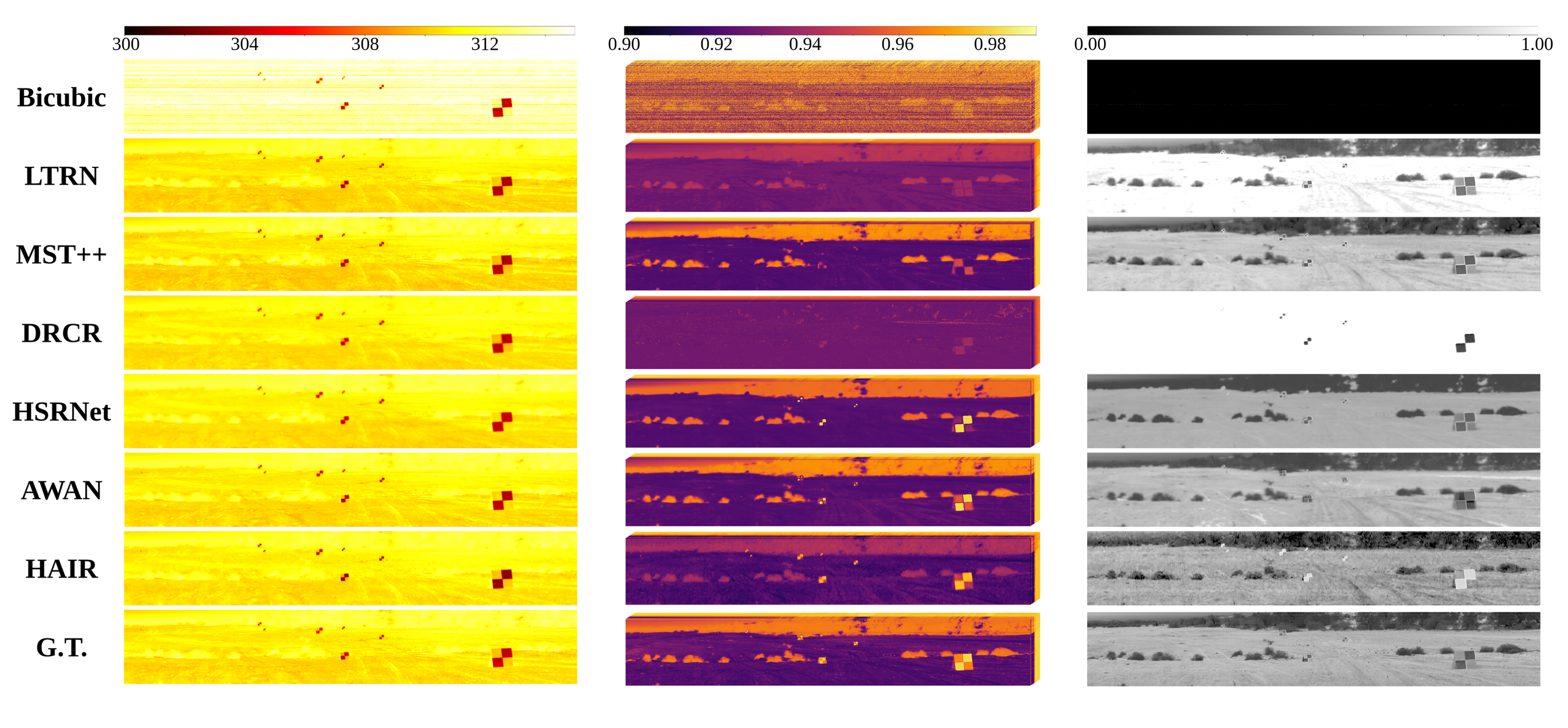}
    \caption{HADAR retrievals in the $8\times$ spectral super-resolution experiment. Columns 1--3 show temperature ($T$, in K), emissivity ($e$), and normalized texture ($X$), respectively.}
    \label{fig:ssr_etc}
\end{figure*}

Figure~\ref{fig:ssr_reflective} shows that existing SSR methods tend to over-smooth or misalign high-frequency atmospheric signatures under severe undersampling, while HAIR preserves them more faithfully. This improves the estimate of $V_{sky}$ and yields a more structured texture $X$ in Fig.~\ref{fig:ssr_etc}.

\subsubsection{Discussion on Practicality and Conclusion}
Unlike paired-data SSR networks, HAIR is training-free and uses the downwelling sky signal as a physical reference, making it less dependent on large HR-LR training sets and more adaptable across various cameras.

\subsection{ROBUSTNESS ANALYSIS}
\label{sec:robustness}

We analyze robustness at both module and system levels, covering noise intensity, stripe severity, catastrophic band corruption, spectral loss, and sensor architecture.

\subsubsection{Robustness in Module A.I}
Module A.I is evaluated by degradation estimation and bad-band exclusion. Table~\ref{tab:module_a_error} shows that $s_{1,k}$ and $s_{2,k}$ estimate Gaussian noise and stripe severity with errors below 1.31\%, indicating reliable band-level characterization. Under the inpainting setting with $\tilde{b}=0.2$, Table~\ref{tab:crop_band} shows that explicitly excluding detected bad bands improves all metrics, confirming the practical value of isolating invalid measurements before reconstruction.
\begin{table}[htbp]
    \scriptsize
    \setlength{\tabcolsep}{2.5pt}
    \renewcommand{\arraystretch}{0.95}
    \caption{Estimation errors of degradation indicators.}
    \label{tab:module_a_error}
    \centering
    \begin{tabular}{c cc cc cc cc}
        \toprule
        \multirow{3}{*}{Index}
                         & \multicolumn{3}{c}{$s_{1,k}$}
                         & \multicolumn{3}{c}{$s_{2,k}$}                                                 \\
        \cmidrule(lr){2-4} \cmidrule(lr){5-7}
                         & MSE                           & MAE    & MAPE (\%)
                         & MSE                           & MAE    & MAPE (\%)                            \\
        \midrule
        Estimation Error & 0.0188                        & 0.0946 & 1.2854    & 0.0022 & 0.0345 & 1.3012 \\
        \bottomrule
    \end{tabular}
\end{table}
\begin{table}[htbp]
    \scriptsize
    \setlength{\tabcolsep}{2.5pt}
    \renewcommand{\arraystretch}{0.95}
    \caption{Ablation study of the corrupted-band processing strategy.}
    \label{tab:crop_band}
    \centering
    \begin{tabular}{c c c c c c}
        \toprule
        Method        & PSNR$\uparrow$   & SSIM$\uparrow$  & ERGAS$\downarrow$ & RMSE$\downarrow$ & SAM$\downarrow$ \\

        \midrule
        HAIR          & \textbf{52.2184} & \textbf{0.9847} & \textbf{0.2935}   & \textbf{0.0296}  & \textbf{0.1166} \\
        HAIR w/o Crop & 50.6103          & 0.9804          & 0.3555            & 0.0364           & 0.1279          \\

        \bottomrule
    \end{tabular}
    \\[2pt]
    \parbox{\linewidth}{\centering \footnotesize
        The best results are highlighted in \textbf{bold}. PSNR is in dB.
    }
\end{table}
\subsubsection{Robustness in Module A.II--A.III}
Module A.II--A.III is tested under varying stripe intensity and Gaussian noise. The cross-comparison in Table~\ref{tab:denoising_comparison_clean} shows stable performance across tested combinations. The destriper is further supported by the derivation and empirical convergence validation in Supplementary Appendix~A; at $\sigma=1.0$ and $\tilde{s}=0.1$, the objective and restored HSI stabilize within 50 iterations. FHyDe~\cite{ZhuangFastHyperspectralImage2018} then serves as a reliable denoiser within the HAIR pipeline.

\subsubsection{Robustness in Modules C--E}
Modules C--E depend on the downwelling reference forward-modeled from atmospheric profiles and ambient temperature via libRadtran. Its generation and validity are provided in Supplementary Appendix~B and Fig.~A2. Under a severe test with 40 consecutive removed bands, Figs.~A3 and A4 in the Supplementary Appendix show that the recovered sky signal remains aligned with the reference and that the HADAR inversion remains stable.

\subsubsection{Overall Robustness of HAIR}
At the system level, Fig.~A5 in the Supplementary Appendix shows that HAIR recovers cleaner $T$-$e$-$X$ representations under heavy noise and severe non-uniform spectral shifts. This supports the combined role of degradation clean-up, physics-driven calibration, and TeX decomposition-synthesis under real sensing conditions.

\subsection{Computational Efficiency and Scalability Analysis}
\begin{figure}
    \centering
    \includegraphics[width=1\linewidth]{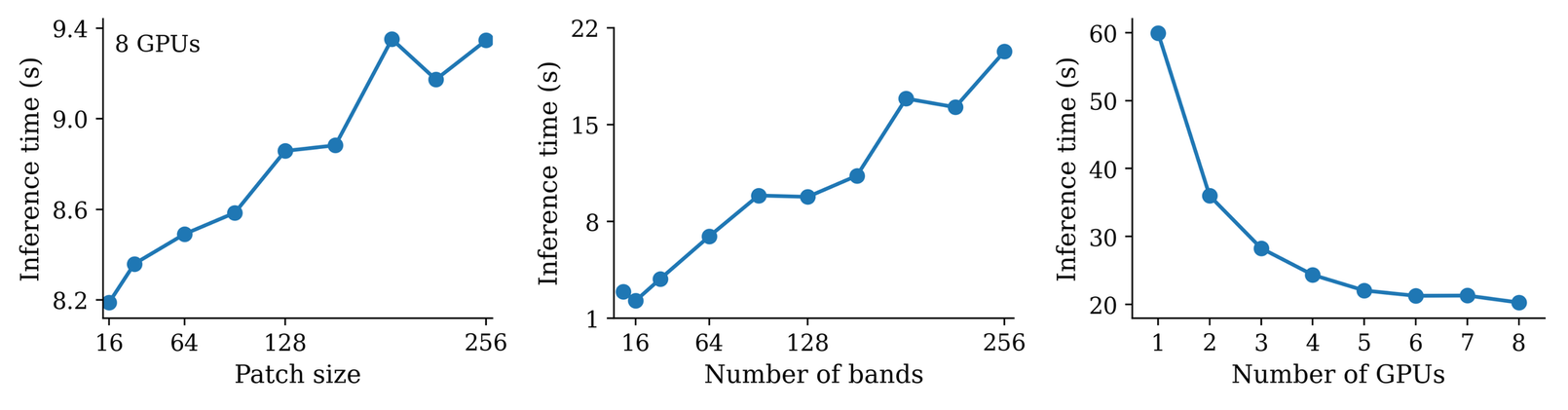}
    \caption{Computational efficiency of HAIR under different configurations. From left to right, the three panels show the inference time as a function of patch size, number of spectral bands, and number of GPUs, respectively.}
    \label{fig:computational_hair_efficiency}
\end{figure}

HAIR remains efficient despite its multi-stage design because the core modules are implemented with parallel JAX. In the efficiency evaluation, CPU-demanding tasks were run on a Threadripper Pro 9985WX, while GPU-demanding tasks were run on an eight-card RTX Pro 6000 platform. Figure~\ref{fig:computational_hair_efficiency} shows near-linear scaling with patch size and spectral dimensionality, and GPU scaling that saturates at about 20~s for a $260\times1500\times256$ cube. Tables~\ref{tab:runtime_restoration} and \ref{tab:time_comparison_breakdown} further show competitive runtime among training-free methods and identify the per-module cost distribution.

\begin{table}[htbp]
    \centering
    \scriptsize
    \setlength{\tabcolsep}{2.5pt}
    \renewcommand{\arraystretch}{0.95}
    \begin{threeparttable}
        \caption{Runtime comparison among training-free methods on denoising tasks.}
        \label{tab:runtime_restoration}
        \begin{tabular}{ c c c c c c c c}
            \toprule
            {Metric}
                   & SSTV   & LRTDTV & E3DTV  & NGMeet & FHyDe             & DIP2d  & HAIR           \\
            \midrule
            Time (s)
                   & 133.33 & 438.76 & 227.07 & 477.94 & \underline{23.56} & 380.78 & \textbf{20.29} \\
            Device & CPU    & CPU    & CPU    & CPU    & CPU               & GPU    & CPU/GPU        \\
            \bottomrule
        \end{tabular}
    \end{threeparttable}
    \\[2pt]
    \parbox{\linewidth}{\centering \footnotesize
        The best and second-best results are highlighted in \textbf{bold} and \underline{underlined}.
    }
\end{table}

\begin{table}[htbp]
    \centering
    \scriptsize
    \setlength{\tabcolsep}{2.5pt}
    \renewcommand{\arraystretch}{0.95}
    \caption{Average inference time of individual modules in HAIR.}
    \label{tab:time_comparison_breakdown}
    \begin{tabular}{l cccccccc}
        \toprule
        Sub-Module    & A.I  & A.II & A.III & B    & C+D  & E+F  & Others & Total \\
        \midrule
        Inference (s) & 0.36 & 2.76 & 8.05  & 0.02 & 0.51 & 5.57 & 3.02   & 20.29 \\
        \bottomrule
    \end{tabular}
\end{table}

\section{CONCLUSION}
In this paper, we present HAIR, a HADAR-based thermal infrared hyperspectral image restoration framework. HAIR establishes a unified degradation formulation for ground-based TIR-HSI, and achieves restoration with three stages: degradation clean-up, physics-driven calibration, and TeX decomposition-synthesis. Built upon sensor-aware recovery, an external downwelling reference, and HRE-constrained reconstruction, HAIR enables physically consistent restoration under spectral undersampling, non-uniform wavelength shifts, and catastrophic band-wise corruption.

Experiments on denoising, inpainting, spectral calibration, and spectral super-resolution show that HAIR achieves consistent gains over representative model-based and deep learning baselines in the evaluated settings. The results also indicate improved visual quality, more plausible TeX retrievals, and favorable runtime scaling under the tested configurations. These findings suggest that HRE-constrained restoration is a promising physics-grounded direction for TIR-HSI recovery and future HADAR-oriented imaging applications.

\bibliographystyle{IEEEtran}
\bibliography{IEEEabrv,references}

\clearpage
\appendices
\setcounter{equation}{0}
\setcounter{figure}{0}
\setcounter{table}{0}
\setcounter{algorithm}{0}
\renewcommand{\theequation}{A\arabic{equation}}
\renewcommand{\thefigure}{A\arabic{figure}}
\renewcommand{\thetable}{A\arabic{table}}
\renewcommand{\thealgorithm}{A\arabic{algorithm}}
\renewcommand{\theHequation}{appendix.A\arabic{equation}}
\renewcommand{\theHfigure}{appendix.A\arabic{figure}}
\renewcommand{\theHtable}{appendix.A\arabic{table}}

\section{Destriper Sub-algorithm}
\label{sec:supp_destriper}
\subsection{ADMM Formulation}
This appendix supplements Module A.II in Section~IV-B2 of the main paper by providing the complete variational formulation, implementation details, and empirical convergence validation of the proposed destriper. Following the main paper, for each valid pushbroom band $k\in\Omega_g$, we explicitly decompose the degraded observation $\mathcal{Y}$ into a non-stripe component $\mathcal{Z}$ and a coherent stripe component $\mathcal{S}$ through the following adaptive variational problem:
\begin{equation}
    \label{eq:destriper_main_supplementary}
    \begin{split}
        \min_{\mathcal{Z},\mathcal{S}} \;
         & \frac{1}{2}\|\mathcal{Y}-\mathcal{Z}-\mathcal{S}\|_F^2
        + \lambda_1\|\nabla_x\mathcal{Z}\|_1 + \lambda_2\|\nabla_y\mathcal{Z}\|_1 \\
         & + \lambda_3\|\nabla_{yy}\mathcal{Z}\|_1
        + \lambda_4\|\nabla_x\mathcal{S}\|_1
        + \lambda_5\|\mathcal{S}\|_1,
    \end{split}
\end{equation}
where $\lambda_2=m s_{2,k}$, in which $m$ is a scaling factor controlling the vertical smoothing and $s_{2,k}$ is the stripe-strength estimate defined in main-paper Eq.~(8), $\|\cdot\|_F$ denotes the Frobenius norm, $\nabla_x$ and $\nabla_y$ are first-order horizontal and vertical difference operators, and $\nabla_{yy}$ is the second-order vertical difference operator.

In our implementation, we set $\lambda_1 = \lambda_3 = \lambda_5 =0.005$, $m = 2$, $\lambda_4 = 1.0$, and use $K=50$ iterations. The penalties are set to $\rho_1 = \rho_2= 0.1$, $\rho_3 = \rho_5 = 0.05$, $\rho_4 = 0.2$.

To solve main-paper Eq.~(12), we adopt the Alternating Direction Method of Multipliers (ADMM). Introducing auxiliary variables $\mathcal{V}_i$ and Lagrange multipliers $\mathcal{M}_i$ for $i=1,\dots,5$, the corresponding augmented Lagrangian is formulated as
\begin{equation}
    \label{eq:alm_destriper}
    \begin{aligned}
        \mathcal{L}_{\rho}(\mathcal{Z}, \mathcal{S}, \{\mathcal{V}_i\}, \{\mathcal{M}_i\})
        = & \,
        \frac{1}{2}\|\mathcal{Y}-\mathcal{Z}-\mathcal{S}\|_F^2
        + \sum_{i=1}^{5}\lambda_i \|\mathcal{V}_i\|_1 \\
          & +
        \sum_{i=1}^{5}\frac{\rho_i}{2}
        \left\|
        \mathcal{D}_i(\mathcal{T})-\mathcal{V}_i+\frac{\mathcal{M}_i}{\rho_i}
        \right\|_F^2,
    \end{aligned}
\end{equation}
where $\mathcal{T}\in\{\mathcal{Z},\mathcal{S}\}$ denotes the target variable, and $\mathcal{D}_i$ denotes the corresponding linear operator (e.g., $\nabla_x$, $\mathbb{I}$). The resulting optimization is decomposed into the following FFT-based primal updates, shrinkage steps, and multiplier updates.

Assuming periodic boundary conditions, the convolution operators $\mathcal{D}_i$ are diagonalizable in the Fourier domain. The updates for $\mathcal{Z}$ and $\mathcal{S}$ can be obtained via Fast Fourier Transform (FFT):
\begin{subequations}
    \label{eq:update_XS}
    \begin{align}
        \mathcal{Z}^{k+1}
         & =
        \mathcal{F}^{-1}\!\left(
        \frac{
            \mathcal{F}(\mathcal{Y}-\mathcal{S}^{k})
            +
            \sum_{i=1}^{3}\rho_i \overline{\hat{\mathcal{D}}_i}\circ \hat{\mathcal{T}}_i^{k}
        }{
            \mathbf{1} + \sum_{i=1}^{3}\rho_i |\hat{\mathcal{D}}_i|^2
        }
        \right), \label{eq:update_X} \\
        \mathcal{S}^{k+1}
         & =
        \mathcal{F}^{-1}\!\left(
        \frac{
            \mathcal{F}(\mathcal{Y}-\mathcal{Z}^{k+1})
            +
            \sum_{j=4}^{5}\rho_j \overline{\hat{\mathcal{D}}_j}\circ \hat{\mathcal{T}}_j^{k}
        }{
            \mathbf{1} + \sum_{j=4}^{5}\rho_j |\hat{\mathcal{D}}_j|^2
        }
        \right), \label{eq:update_S}
    \end{align}
\end{subequations}
where $\mathcal{F}$ denotes the 2D FFT, $\overline{(\cdot)}$ denotes complex conjugation, $\hat{\mathcal{D}}_i=\mathcal{F}(\mathcal{D}_i)$, $\hat{\mathcal{T}}_i^k:=\mathcal{F}(\mathcal{V}_i^k-\mathcal{M}_i^k/\rho_i)$ and all multiplications and divisions are element-wise.

The auxiliary variables $\{\mathcal{V}_i\}_{i=1}^{5}$ admit closed-form proximal updates via soft-thresholding:
\begin{equation}
    \label{eq:update_V}
    \mathcal{V}_i^{k+1}
    =
    \mathrm{soft}\!\left(
    \mathcal{D}_i(\mathcal{T}^{k+1}) + \frac{\mathcal{M}_i^k}{\rho_i},
    \frac{\lambda_i}{\rho_i}
    \right),
\end{equation}
where $ \mathrm{soft}(x,\tau)=\mathrm{sign}(x)\cdot\max(|x|-\tau,0)$.

Then, the Lagrange multipliers are updated by
\begin{equation}
    \label{eq:update_alm_multipliers}
    \mathcal{M}_i^{k+1}
    =
    \mathcal{M}_i^k
    +
    \rho_i\bigl(\mathcal{D}_i(\mathcal{T}^{k+1})-\mathcal{V}_i^{k+1}\bigr).
\end{equation}

The complete procedure is summarized in Algorithm~\ref{alg:destriper_total}.

\begin{algorithm}[htbp]
    \caption{Destriper Sub-algorithm}
    \label{alg:destriper_total}
    \begin{algorithmic}[1]
        \REQUIRE Noisy HSI $\mathcal{Y}$, regularization parameters $\{\lambda_i\}_{i=1}^{5}$, penalty parameters $\{\rho_i\}_{i=1}^{5}$, maximum iterations $K$.
        \ENSURE Non-stripe image $\mathcal{Z}$ and stripe component $\mathcal{S}$.
        \STATE \textbf{Initialize:} $\mathcal{Z}^0=\mathcal{Y}$, $\mathcal{S}^0=\mathbf{0}$, and $\{\mathcal{V}_i^0,\mathcal{M}_i^0\}_{i=1}^{5}=\mathbf{0}$.
        \STATE \textbf{Pre-compute:} $\hat{\mathcal{D}}_i$ and the denominators in Eq.~\eqref{eq:update_XS}.
        \FOR{$k=0$ to $K-1$}
        \STATE Update $\mathcal{Z}^{k+1}$ via Eq.~\eqref{eq:update_X}.
        \STATE Update $\mathcal{S}^{k+1}$ via Eq.~\eqref{eq:update_S}.
        \STATE Update $\{\mathcal{V}_i^{k+1}\}_{i=1}^{5}$ via Eq.~\eqref{eq:update_V}.
        \STATE Update $\{\mathcal{M}_i^{k+1}\}_{i=1}^{5}$ via Eq.~\eqref{eq:update_alm_multipliers}.
        \ENDFOR
        \RETURN $\mathcal{Z}^{K}, \mathcal{S}^{K}$
    \end{algorithmic}
\end{algorithm}

\subsection{Convexity and Empirical Convergence Validation}
The destriping objective in main-paper Eq.~(12) is well posed because its data-fidelity, image-prior, and stripe-prior terms are convex, and the auxiliary constraints introduced for ADMM are linear. In particular, the data-fidelity term is quadratic, while the image prior and stripe prior are composed of proper closed convex TV/$\ell_1$ regularizers rather than non-convex projection operators~\citesupp{convexity_provement_Wright-Ma-2022}. This convex formulation makes the subproblems numerically stable; however, because the implementation uses alternating multi-block updates, we do not claim a universal convergence theorem for the complete solver.

\begin{figure}[htbp]
    \centering
    \includegraphics[width=1\linewidth]{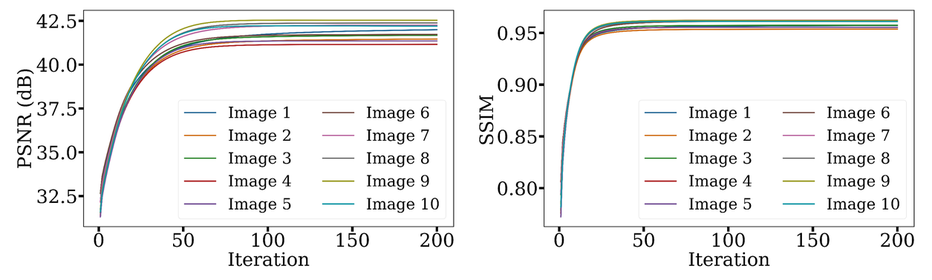}
    \caption{Empirical convergence curve of the proposed destriper under the simulated denoising setting $\sigma^2=1.0,\tilde{s}=0.1$.}
    \label{fig:convergence_von_destriper}
\end{figure}

We therefore validate the solver behavior empirically under the simulated denoising setting used in the main paper, namely $\sigma^2=1.0$ and $\tilde{s}=0.1$. As shown in Fig.~\ref{fig:convergence_von_destriper}, the objective value decreases monotonically and the reconstructed HSI rapidly stabilizes within the fixed budget of 50 iterations. This evidence supports the numerical stability of the proposed destriper under relatively strong stripe and Gaussian interference.

\section{Validity and Robustness of Modules C--E}
\label{sec:supplementary_validity_and_robustness_of_module_c_e}
\subsection{Reliability of the Forward-Modeled Reference}
\label{sec:supplementary_reliability_of_the_forward_modeled_reference}
The reliability of the forward-modeled downwelling reference stems from the fact that it is generated from first-principles radiative transfer physics. The distribution of photons in a dilute gas is governed by the Boltzmann equation~\citesupp{chandrasekhar2013radiative,liou2002introduction}
\begin{equation}
    \label{eq:BoltzmannEquation}
    \frac{\partial f}{\partial t}
    + \nabla_{\mathbf{r}} ( \mathbf{v} f )
    + \nabla_{\mathbf{p}} ( \mathbf{F} f )
    = Q(\mathbf{r}, \hat{n}, \nu, t),
\end{equation}
where $f$ denotes the photon distribution function. Neglecting relativistic effects and assuming constant photon velocity between collisions, Eq.~\eqref{eq:BoltzmannEquation} reduces, under steady-state and plane-parallel assumptions, to the one-dimensional radiative transfer equation
\begin{equation}
    \label{eq:1DRTE}
    \mu \frac{\partial I}{\partial z}
    =
    -\beta^{\mathrm{ext}}(z,\nu)\, I
    + \mathcal{J}
    + \beta^{\mathrm{abs}}(z,\nu)\, B[T(z)],
\end{equation}
where $I \equiv I(z,\mu,\phi,\nu)$ is the spectral radiance, and the scattering contribution is
\begin{equation}
    \label{eq:scattering}
    \begin{aligned}
        \mathcal{J}
         & =
        \frac{1}{4\pi}
        \int_{0}^{\infty} d\nu'\,
        \beta^{\mathrm{sca}}(z,\nu,\nu')\int_{0}^{2\pi} d\phi' \\
         & \quad \times
        \int_{-1}^{1} d\mu'\,
        p(z,\mu,\phi;\mu',\phi')\,
        I(z,\mu',\phi',\nu').
    \end{aligned}
\end{equation}
The absorption, scattering, and extinction coefficients are given by
\begin{equation}
    \label{eq:coefficients}
    \begin{aligned}
        \beta^{\mathrm{abs}}(z,\nu) & = \sum_i n_i(z)\, \sigma_i^{\mathrm{abs}}(\nu),              \\
        \beta^{\mathrm{sca}}(z,\nu) & = \sum_i n_i(z)\, \sigma_i^{\mathrm{sca}}(\nu),              \\
        \beta^{\mathrm{ext}}(z,\nu) & = \beta^{\mathrm{abs}}(z,\nu) + \beta^{\mathrm{sca}}(z,\nu).
    \end{aligned}
\end{equation}

In the thermal infrared regime, molecular scattering is typically negligible because Rayleigh scattering decays rapidly with wavelength ($\propto \lambda^{-4}$)~\citesupp{pierrehumbert2010principles}. As a result, the downwelling radiance is governed by molecular absorption and thermal emission, which makes the forward model particularly well grounded in the TIR band. Under this regime, the key quantities required for simulation are the atmospheric temperature profile $T(z)$, the molecular number-density profiles $n_i(z)$, and the spectroscopic absorption cross sections $\sigma_i^{\mathrm{abs}}(\nu)$~\citesupp{GORDON2026109807}, together with boundary conditions.

\begin{figure}
    \centering
    \includegraphics[width=1\linewidth]{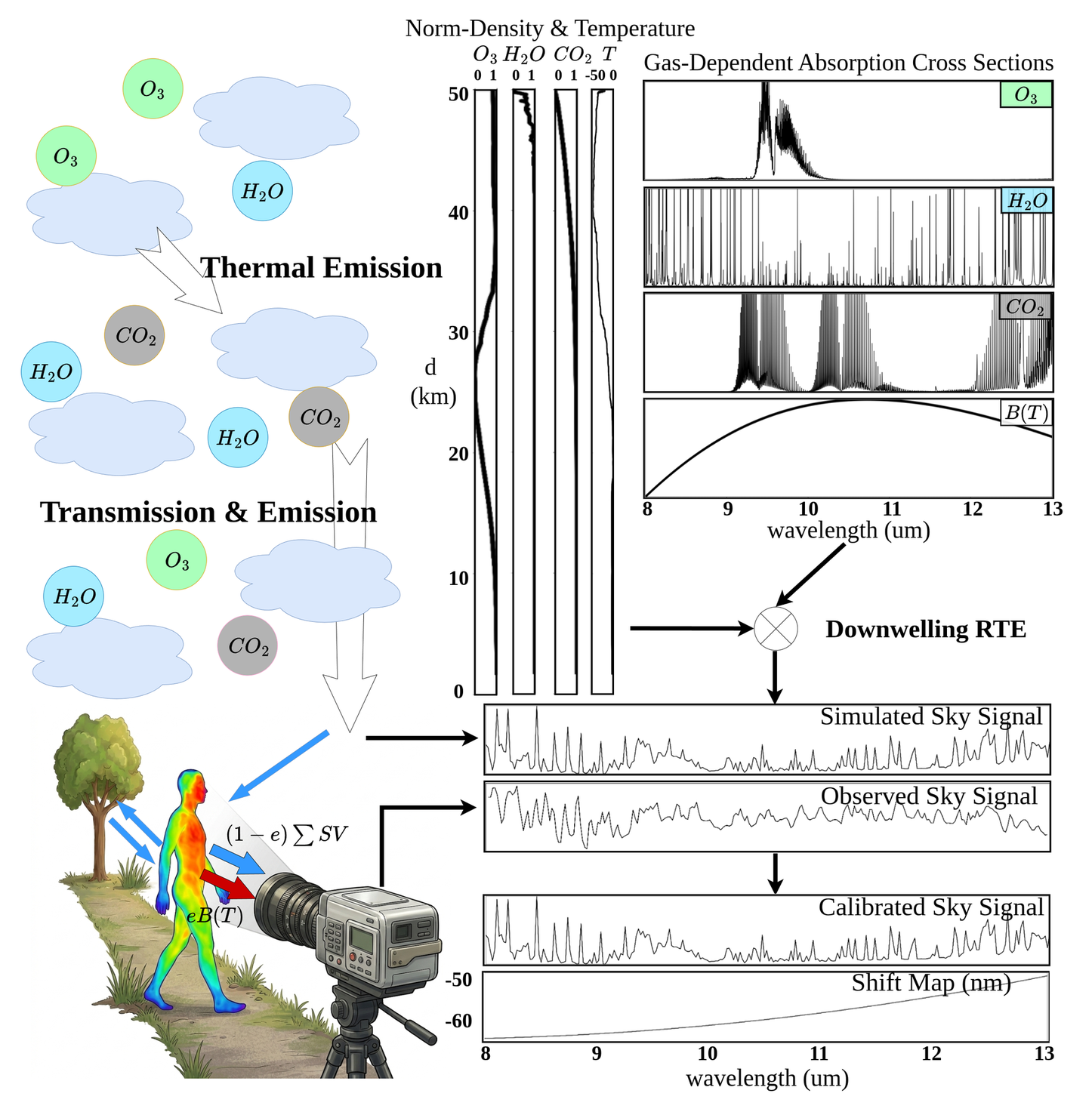}
    \caption{Overview of the spectral calibration process (Module C-E). A simulated downwelling signal $\mathbf{s}_{\mathrm{r}}$ is generated from atmospheric emission and transmission under the downwelling radiative transfer equation in Eq.~\eqref{eq:1DRTE} via libRadtran~\protect\citesupp{libradtran_software_emde2016libradtran, GORDON2026109807}. The observed sky signal $\mathbf{s}$ is extracted from the measurement, and calibration is performed via main-paper Eq.~(19), yielding the shift map $\Delta\lambda_k$ (main-paper Eq.~(20)) and the calibrated sky signal $\hat{\mathbf{s}}$ (main-paper Eq.~(18)).}
    \label{fig:libradtran_forwarded}
\end{figure}

Given these physically measurable inputs, the downwelling sky radiance is obtained by numerically solving Eq.~\eqref{eq:1DRTE} using the DISORT solver~\citesupp{disort} implemented in libRadtran~\citesupp{libradtran_software_emde2016libradtran}. Therefore, the resulting reference is not an empirical template, but a physically grounded estimate determined by atmospheric state variables and molecular spectroscopy. This physical derivation is precisely what makes the simulated downwelling signal a reliable reference for the subsequent spectral calibration, as illustrated in Fig.~\ref{fig:libradtran_forwarded}.

\subsection{Robustness of Spectral Calibration}
\label{sec:supplementary_robustness_of_spectral_calibration}
Robust spectral calibration is particularly important in TIR HSI because catastrophic spectral loss is often not randomly distributed, but tends to concentrate near spectral extremities. In FTIR systems, such degradation frequently arises from the rapid decline of beamsplitter transmittance and detector quantum efficiency near the edge bands~\citesupp{FTIR_noise_analysis_optical_engineering_2012noise}, while in pushbroom systems severe band failure may also accumulate at specific spectral rows due to sensor non-uniformity and hardware instability~\citesupp{pushbroom_flicker_noise_scribner1988physical}. As a result, long contiguous missing intervals near the spectral boundaries are a realistic and practically important failure mode in TIR imaging.

\begin{figure}
    \centering
    \includegraphics[width=1\linewidth]{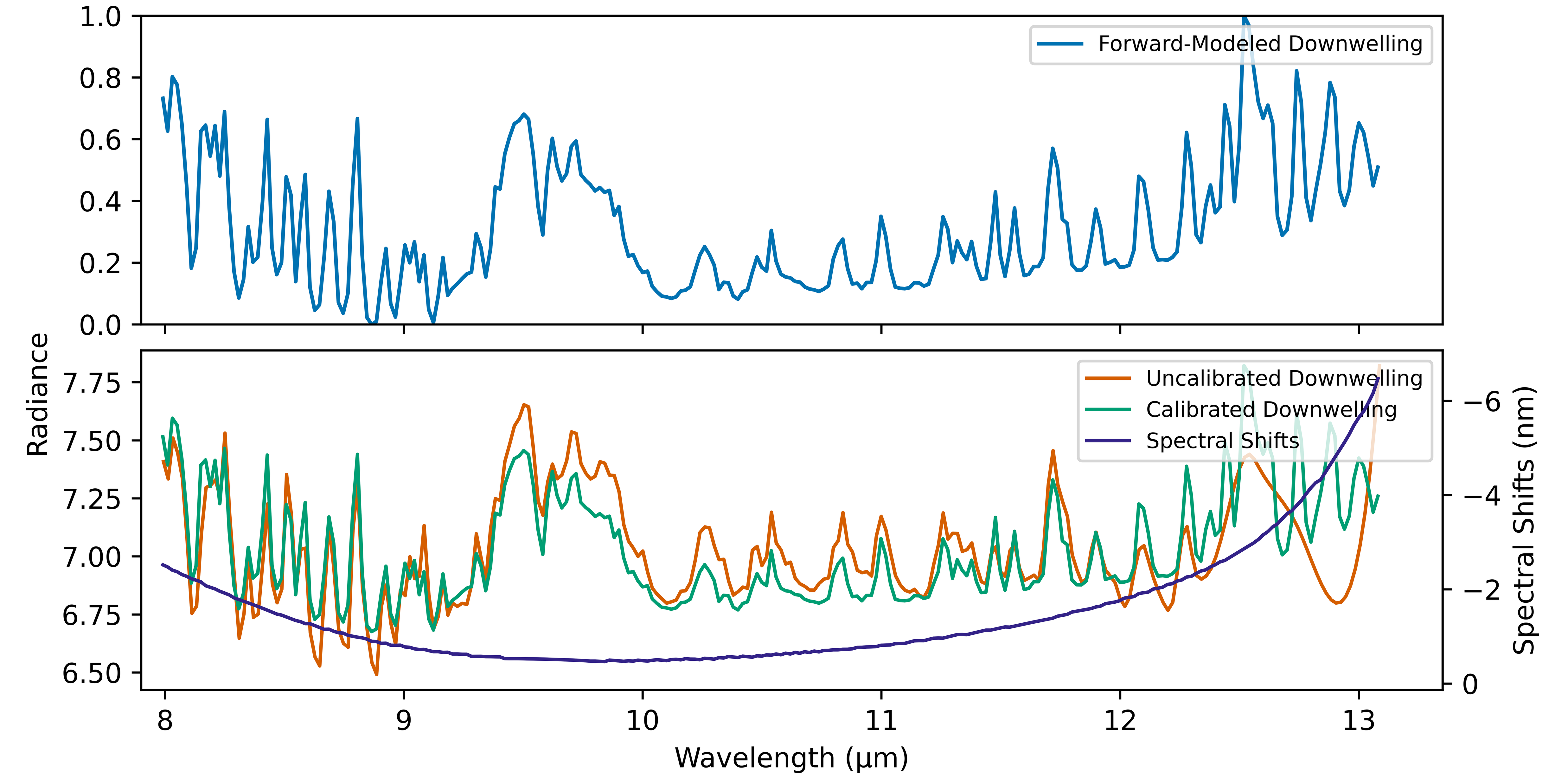}
    \caption{Spectral calibration under severe spectral loss. The uncalibrated observed sky signal $\mathbf{s}$, reconstructed after removing 40 consecutive bands, is aligned against the reference $\mathbf{s}_{\mathrm{r}}$ (main-paper Eq.~(19)) to estimate the shift map $\Delta\lambda_k$ (main-paper Eq.~(20)) and the calibrated sky signal $\hat{\mathbf{s}}(k)$ (main-paper Eq.~(18)).}
    \label{fig:severe_condition_calibration}
\end{figure}

\begin{figure}
    \centering
    \includegraphics[width=1\linewidth]{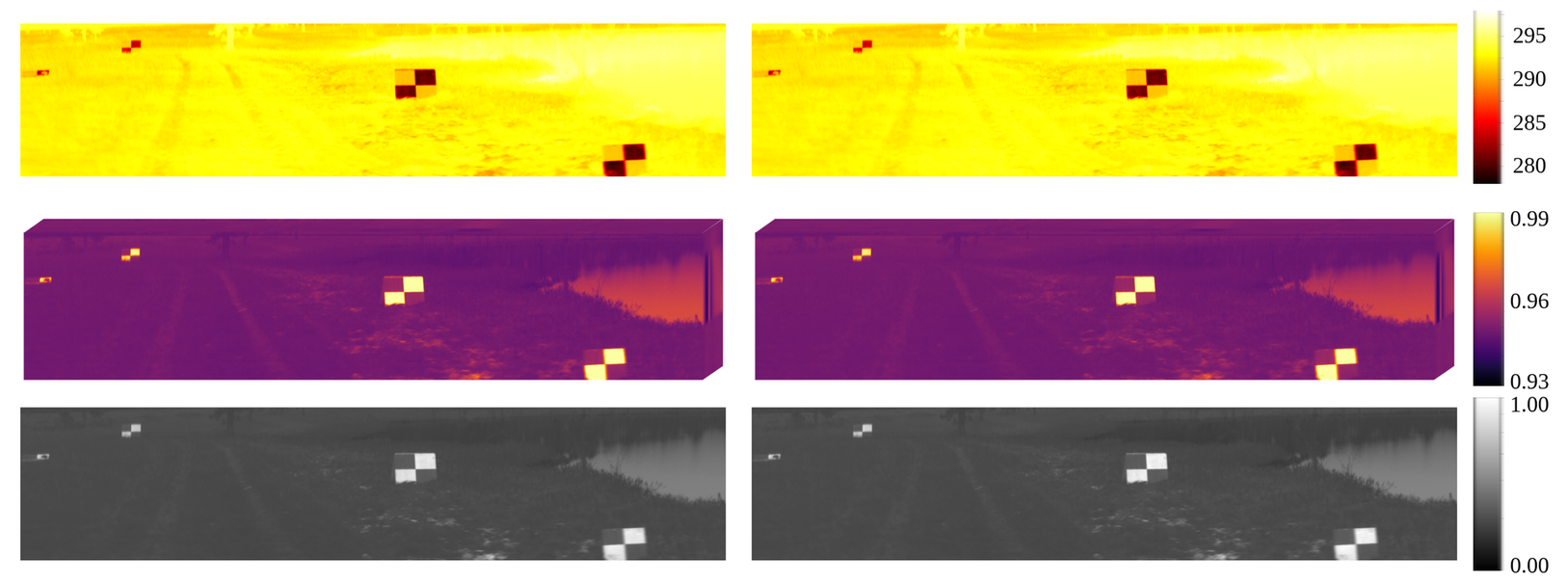}
    \caption{HADAR retrieval comparison on uncalibrated and calibrated HSI under severe contiguous missing-band corruption. Left: Uncalibrated. Right: Calibrated. Rows 1--3 show temperature ($T$ in K), emissivity ($e$), and normalized texture ($X$), respectively.}
    \label{fig:band_correction_severe_etx}
\end{figure}

To evaluate HAIR under this extreme setting, we select one HSI cube from the DARPA dataset~\citesupp{YellinConcurrentBandSelection2024} with size $260\times1500\times256$, and remove the last 40 consecutive corrupted bands. In this case, interpolating the extracted atmospheric signature $\mathbf{s}' \in \mathbb{R}^{|\Omega_g|}$ to obtain the observed sky signal $\mathbf{s}$ via main-paper Eq.~(16) introduces severe extrapolation error, as shown in Fig.~\ref{fig:severe_condition_calibration}. Despite this substantial deviation, the proposed spectral calibration remains effective. As shown in Figs.~\ref{fig:severe_condition_calibration} and \ref{fig:band_correction_severe_etx}, our method still recovers a physically reasonable shift map $\Delta\lambda_k$ and calibrated sky signal $\hat{\mathbf{s}}$, and the HADAR inversion remains stable with preserved thermodynamic consistency. These results confirm that the proposed strategy is robust even under contiguous spectral loss.

\section{Robustness of HAIR}
\label{sec:supplementary_robustness_of_HAIR}
\begin{figure*}
    \centering
    \includegraphics[width=1\linewidth]{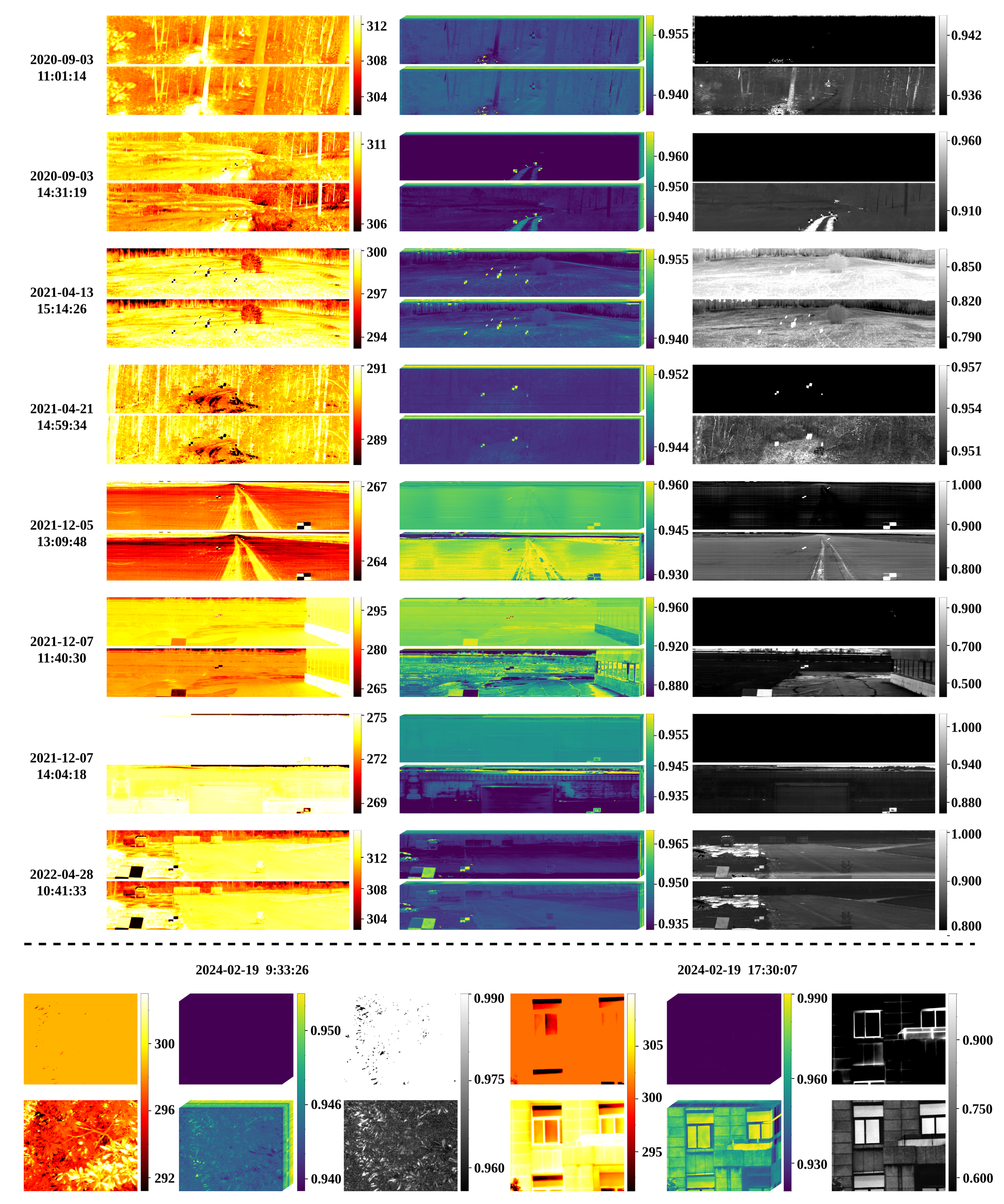}
    \caption{Robustness of HAIR under diverse real-world conditions. The upper part shows examples from the DARPA dataset, and the lower part shows examples from in-lab FTIR measurements, covering multiple scene categories, object types, and acquisition times. In each case, the upper image corresponds to the retrievals from the degraded input, while the lower image shows the HAIR-restored result. From left to right, the three columns correspond to temperature $T$ (in K), emissivity $e$, and normalized texture $X$. The time labels indicate the local acquisition time.}
    \label{fig:final_robustness_von_hair}
\end{figure*}

The main paper presents a few representative restoration examples (Figs.~5 and 6). To further demonstrate the robustness and decisive role of HAIR, we provide additional results on the DARPA dataset~\citesupp{YellinConcurrentBandSelection2024} and in-lab FTIR measurements, covering different scene categories, object types, and acquisition times. As shown in Fig.~\ref{fig:final_robustness_von_hair}, the proposed framework transforms noisy TIR HSI, suffering from severe TeX Degeneracy due to high interference~\citesupp{BaoHeatassistedDetectionRanging2023,xu2026universalcomputationalthermalimaging}, into a physically interpretable $T$-$e$-$X$ representation. The improvement is particularly evident in emissivity $e$ and normalized texture $X$, where HAIR recovers clearer semantic structure and more realistic details across both pushbroom and FTIR data. These results further confirm the robustness of the proposed three-stage framework under diverse real-world sensing conditions.

\bibliographystylesupp{IEEEtran}
\bibliographysupp{references}

\end{document}